\documentclass{article}

\usepackage[final]{corl_2020} 

\usepackage[table]{xcolor}
\usepackage[utf8]{inputenc}
\usepackage[T1]{fontenc}    
\usepackage{hyperref}       
\usepackage{comment}
\usepackage{graphicx}
\usepackage{graphics}
\usepackage{float}
\usepackage{graphicx}
\usepackage{amssymb}
\usepackage{amsmath}
\usepackage{caption}

\usepackage{fancyvrb}
\usepackage{amssymb}
\usepackage{amsmath}
\usepackage{amsfonts}
\usepackage{amssymb}
\usepackage{array}
\usepackage{subfigure}
\usepackage{ragged2e}
\usepackage{tabu}
\usepackage{balance}
\usepackage{algorithmicx}
\usepackage{algorithm}
\usepackage{algpseudocode}
\usepackage{graphicx}
\usepackage{float}
\usepackage{hyperref}
\usepackage{url}
\usepackage{adjustbox}

\usepackage{url}            
\usepackage{booktabs}       
\usepackage{amsfonts}       
\usepackage{nicefrac}       
\usepackage{microtype}      
\usepackage{amsmath}
\usepackage{multirow}
\usepackage{graphicx} 
\usepackage{graphics}
\usepackage{multirow}
\usepackage{float}
\usepackage{fancyvrb}
\usepackage{amssymb}
\usepackage{amsmath}
\usepackage{amsfonts}
\usepackage{amssymb}
\usepackage{array}
\usepackage{ragged2e}
\usepackage{tabu}
\usepackage{algorithmicx}
\usepackage{algorithm}
\usepackage{algpseudocode}
\usepackage{todonotes}
\usepackage{url}
\usepackage{booktabs}

\usepackage{adjustbox}
\usepackage{makecell}
\usepackage{amsthm}
\usepackage{comment}

\newtheorem{theorem}{Theorem}[section]

\theoremstyle{definition}
\newtheorem{definition}[theorem]{Definition}

\theoremstyle{remark}

\usepackage{enumitem}

\newcommand{\watermark}{{\mathsf{Watermark}}}
\newcommand{\detect}{{\mathsf{Detect}}}
\newcommand{\aux}{{\mathsf{aux}}}

\def\R{\mathbb{R}}

\def\cX{\mathcal{X}}

\newcommand{\takeaway}[2][]{\todo[inline,linecolor=black,backgroundcolor=orange!10,bordercolor=black,#1]{Takeaway: #2}}

\title{Evaluating Durability: Benchmark Insights into \\ Multimodal Watermarking}

\author{%
  Jielin Qiu$^{1,2}$\footnotemark[1]~~\footnotemark[2]~, William Han$^{2}$\footnotemark[2]~, Xuandong Zhao$^{3}$, 
  \textbf{Shangbang Long}$^{1}$, \\
  \textbf{Christos Faloutsos}$^{2}$, \textbf{Lei Li}$^{2}$ \\
  $^1$Google Research, $^2$Carnegie Mellon University, $^3$University of California, Santa Barbara \\
  {\tt\footnotesize \{jielinqiu,longshangbang\}@google.com, \{wjhan,christos,leili\}@andrew.cmu.edu}
}

\begin{document}
\maketitle
\begin{abstract}
  With the development of large models, watermarks are increasingly employed to assert copyright, verify authenticity, or monitor content distribution. As applications become more multimodal, the utility of watermarking techniques becomes even more critical. The effectiveness and reliability of these watermarks largely depend on their robustness to various disturbances. However, the robustness of these watermarks in real-world scenarios, particularly under perturbations and corruption, is not well understood. To highlight the significance of robustness in watermarking techniques, our study evaluated the robustness of watermarked content generated by image and text generation models against common real-world image corruptions and text perturbations. Our results could pave the way for the development of more robust watermarking techniques in the future. Our project website can be found at \url{https://mmwatermark-robustness.github.io/}. 
\end{abstract}

\section{Introduction}

Multimodal watermarks represent a sophisticated method of embedding information within digital content across various modes, such as audio, video, and images. These watermarks are designed to be imperceptible, or nearly so, to the human senses, yet detectable by specialized software or algorithms. The primary purpose of a watermark is to assert copyright or verify the authenticity of the content. Unlike traditional watermarks, which are limited to a single type of media, multimodal watermarks can be integrated across different formats, enhancing security and flexibility. The robustness of these watermarks against various forms of manipulation and their ability to remain intact even when the content is transformed or compressed is a key aspect of their design, making them essential in copyright protection and digital rights management.

The robustness of multimodal watermarks under various perturbations is a critical aspect of their effectiveness and reliability. In the digital realm, image content often undergoes a variety of transformations, such as compression, scaling, cropping, or format conversion, while textual content often undergoes synonym replacement, paraphrasing, or typing differences, which can potentially alter or obliterate embedded watermarks. The robustness of a watermark to withstand these perturbations is vital to ensure that the embedded data remains intact and retrievable. This is particularly important for copyright protection, piracy detection, and the verification of the authenticity of digital media. A robust watermark ensures that ownership rights are preserved and content integrity is maintained, even when the media is shared across different platforms and undergoes multiple alterations. Furthermore, in sensitive applications such as legal documentation or secure communications, the persistence of a watermark through various distortions is crucial for maintaining the trust and reliability of the information contained within the media. Therefore, understanding the watermark's robustness to perturbations is a key focus.

To our best knowledge, there is currently no comprehensive study of how the perturbations can affect the performance of multimodal watermarks.
Hence, in this work:
\begin{itemize}[leftmargin=*, itemsep=0pt, topsep=0pt]
    \item We evaluate watermark robustness under multimodal perturbations by analyzing 4 image watermarking methods and 4 text watermarking methods. Our study tested the performance of 8 image-to-text models and 8 text-to-image models against 100 image perturbations and 63 text perturbation methods. 
    \item We find that multimodal watermarks are sensitive to distribution shifts caused by image and text perturbations. Specifically, for image perturbations, Zoom Blur consistently shows the highest impact, while Glass Blur is the least harmful one. For text perturbations, Casual consistently shows the highest impact, while OCR is the least harmful. Under image perturbations, SSL-WM seems more stable; while under text perturbations, KGW-WM seems more stable.
    \item We have publicly released our codebase at~\url{https://mmwatermark-robustness.github.io/} with CC BY-NC-SA License.
\end{itemize}

\section{Related Work}

\paragraph{Text Watermarking} 
has become increasingly relevant due to the usage of language models (LMs) for text generation \cite{liu2024survey}. Text watermarking frameworks should integrate seamlessly into a model with minimal impact on the generated text, without altering the model's parameters \cite{kirchenbauer2023watermark, aaronson, christ2023undetectable}. 
Additionally, text watermarks must be robust against distribution shifts, leading to the proposal of a watermarking algorithm that assigns a sequence of arbitrary numbers generated by a random watermark key to a sample from the LM \cite{Kuditipudi2023RobustDW}. However, relying solely on empirical methods to assess the effectiveness of proposed watermarking algorithms is insufficient. In response, a theoretical framework to quantify the robustness of text watermarks was introduced. From this theoretical analysis, an enhanced framework that utilizes a fixed grouping strategy was proposed \cite{zhao2023provable}.
However, these works lack comprehensive evaluations of their watermarking systems under a variety of perturbations.

\vspace{-5pt}
\paragraph{Image Watermarking} 
studies protecting intellectual image property~\cite{cox2007digital}. Recently, encoder/decoder models have been introduced~\cite{ahmadi2020redmark,lee2020resolution,luo2020distortion,zhang2020udh,zhu2018hidden,fernandez2022sslwatermarking,kishore2021fixed,vukotic2018deep}, which have shown promising results in terms of robustness against a broad array of transformations. In the realm of generative models, there have been attempts to watermark the training datasets used to train these models~\cite{yu2021artificial}, an approach that is markedly inefficient as embedding each new message necessitates a separate training pipeline. A more contemporary strategy involves integrating the watermarking process directly with the generative process~\cite{fei2022supervised, lin2022cycleganwm, nie2023attributing, qiao2023novel, wu2020watermarking, yu2022responsible, zhang2020model}, aligning it more closely with the broader literature on model watermarking~\cite{uchida2017embedding}.


\vspace{-5pt}
\paragraph{Robustness of Image or Text Watermarks} 
\cite{An2024BenchmarkingTR} examined the vulnerabilities in various image watermarking techniques. \cite{Zhao2023InvisibleIW} and \cite{Saberi2023RobustnessOA} found that methods like noising and denoising through diffusion models can effectively remove some watermarks. \cite{Jiang2023EvadingWB} studied the robustness of AI-generated content detection that relies on watermarking. \cite{Mofayezi2023BenchmarkingRT} explored the impact of text-guided corruptions on image classifiers. \cite{Kirchenbauer2023OnTR} optimized the watermark generation and detection pipeline for greater reliability in real-world scenarios. Our study aims to challenge both text and image watermarking systems using multimodal perturbations to identify potential vulnerabilities and robustness.

\vspace{-5pt}
\paragraph{Robustness of Multimodal Models}  

For the robustness evaluation of multimodal image-text models under distribution shift, previous works \cite{goh2021multimodal,daras2022discovering,UnderstandingCLIP,StanislavPixels,goh2021multimodal,Noever2021ReadingIB} have tested some pre-trained models, such as CLIP \cite{Radford2021LearningTV}, by attacking them with text patches and adversarial pixel perturbations. Notably, \cite{daras2022discovering} discovered that DALLE-2 \cite{Ramesh2022HierarchicalTI} possesses a hidden vocabulary that enables image generation from absurd prompts, questioning the robustness of its output. \cite{Fang2022DataDD} attributed robustness gains primarily to diverse training distributions. \cite{Cho2022DALLEvalPT} explored the robustness of text-to-image generative models concerning visual reasoning capabilities and social biases. For benchmarking robustness, \cite{Li2021AdversarialVA} compiled an Adversarial VQA dataset to assess the robustness of VQA models. \cite{ChantrySchiappa2022MultimodalRA} examined the robustness of video-text models under perturbations, focusing solely on a video-text retrieval task. Furthermore, \cite{qiu2023benchmarking} investigated the robustness of image-text models to perturbations in both modalities across five downstream tasks, while \cite{Chen2023BenchmarkingRO} evaluated the robustness of adaptation methods across vision-language datasets under multimodal corruptions.

\begin{figure*}[t]
  \centering
  \includegraphics[width=0.95\linewidth]{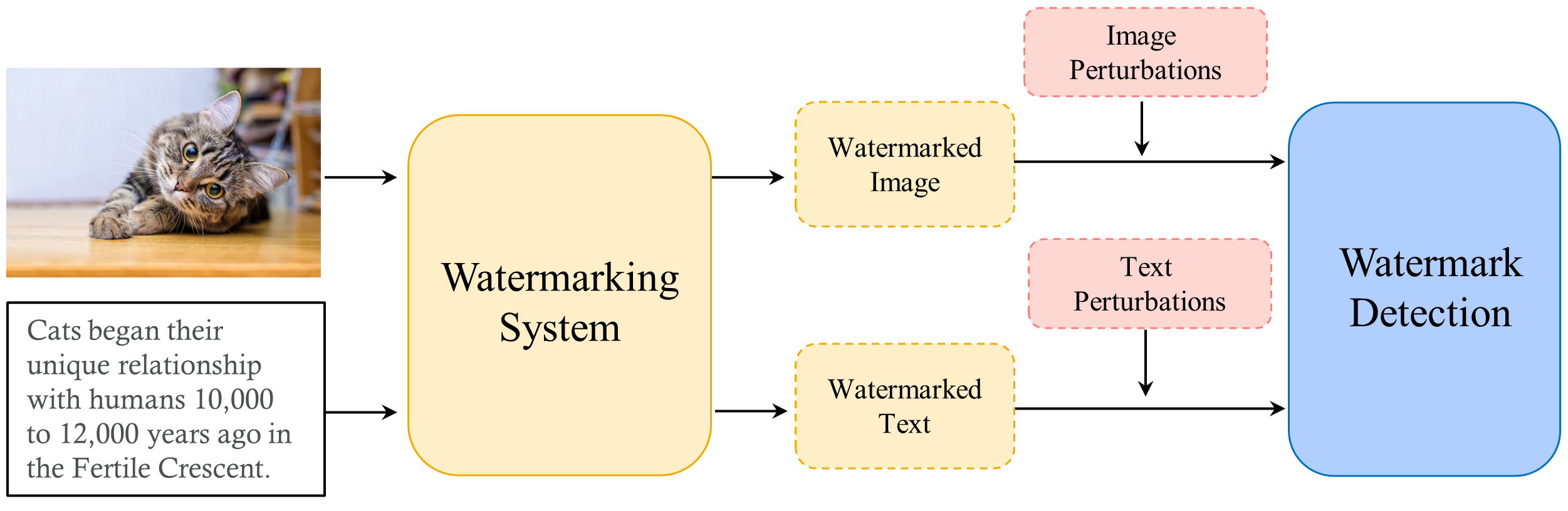}
  \caption{The overall pipeline of our watermarking robustness study. We add watermarks to the generated content and evaluate their robustness under image corruptions and text perturbations. }
  \label{Fig:pipeline}
  \vspace{-10pt}
\end{figure*}

\section{Preliminary: Invisible Watermark Detection}

Before delving into methods for measuring the robustness of watermarks, it is essential to first understand the fundamentals of invisible watermarks and the techniques employed to detect them.

\begin{definition}[Invisible watermark]\label{def
}
Let $x\in \cX$ represent the original content, and let $x_w=\watermark(x,\aux)$ denote the watermarked content, where the watermarking scheme is a function of $x$ and any auxiliary information $\aux$, such as a secret key \cite{Zhao2023InvisibleIW}. A watermark is defined as $\Delta$-\emph{invisible} on a clean image $x$ with respect to a ``distance" function $\mathsf{dist}:\cX \times \cX \rightarrow \R_+$, if $\mathsf{dist}(x,x_w) \leq \Delta$.
\end{definition}

\begin{definition}[Watermark detection]\label{def
}
A watermark detection algorithm $\detect:\cX \times \aux\rightarrow \{0,1\}$ is designed to determine whether a content $\tilde{x}\in\cX$ is watermarked, using auxiliary information such as a secret key ($\aux$) \cite{Zhao2023InvisibleIW}. The algorithm $\detect$ is subject to two types of errors: false positives, where unwatermarked content is incorrectly classified as watermarked, and false negatives, where watermarked content is incorrectly classified as unwatermarked. We define the content $\tilde{x}$ as being drawn from either the null distribution $P_0$ or the watermarked distribution $P_1$. The Type I error, or false positive rate, is denoted as $\epsilon_1:=\Pr_{x\sim P_0}[\detect(x)=1]$, and the Type II error, or false negative rate, is denoted as $\epsilon_2:= \Pr_{x\sim P_1}[\detect(x)=0]$.
\end{definition}

A watermarking scheme is typically engineered so that the distribution of watermarked content, $P_1$, is distinct from that of unwatermarked content, $P_0$. This distinction allows the corresponding detection algorithm, which is carefully designed, to almost perfectly differentiate between the two, aiming to ensure that both Type I error ($\epsilon_1$) and Type II error ($\epsilon_2$) are nearly zero. An attack on a watermarking scheme typically involves post-processing a possibly watermarked image in a way that alters both $P_0$ and $P_1$, with the goal of simultaneously increasing both the Type I and Type II errors, thereby evading detection.

\section{Perturbation Methods}

To evaluate the robustness of watermarks under image and text perturbations, we build a comprehensive evaluation benchmark via perturbing the watermarked, generated text or image.

\vspace{-5pt}
\paragraph{Image Perturbation.}
To simulate real-world corruptions in image data, we employ perturbation strategies adopted from \cite{Hendrycks2019BenchmarkingNN,qiu2023benchmarking,Zhang2024BenchmarkingLM}. These perturbations are categorized into five groups: \textbf{noise, blur, weather, digital}, and \textbf{geometric}. Specifically, we utilize $20$ different image perturbation techniques across these categories: (1) Noise—Gaussian noise, shot noise, impulse noise, speckle noise; (2) Blur—defocus blur, frosted glass blur, motion blur, zoom blur; (3) Weather—snow, frost, fog, brightness; (4) Digital—contrast, elastic transformation, pixelation, JPEG compression; and (5) Geometric—scaling, rotation, shearing, piecewise affine transformation. Acknowledging that real-world corruptions vary in intensity, we introduce variations for each type of corruption as suggested in \cite{Hendrycks2019BenchmarkingNN,Geirhos2019ImageNettrainedCA,michaelis2019dragon}. In our evaluation, each category features 5 severity levels, culminating in a total of $100$ perturbation methods. These strategies, commonly regarded as synthetic distribution shifts, provide a well-defined and manageable starting point. A detailed description of each perturbation method is provided in Table~\ref{table:image_perturbation} in Appendix~\ref{sec:appendix-perturbations}.

\vspace{-5pt}
\paragraph{Text Perturbation.}

To simulate distribution shifts in language data, we have designed 19 text perturbation techniques organized into three categories: \textbf{character-level, word-level}, and \textbf{sentence-level} \cite{qiu2023benchmarking}.
For character-level perturbations, we adopt six strategies from \cite{ma2019nlpaug,qiu2023benchmarking} that simulate common typing errors: keyboard typos, OCR errors, character insertion (CI), character replacement (CR), character swap (CS), and character deletion (CD).
At the word-level, five strategies from EDA and AEDA \cite{Wei2019EDAED,Karimi2021AEDAAE} are used: synonym replacement (SR), word insertion (WR), word swap (WS), word deletion (WD), and punctuation insertion (IP). These techniques reflect various editorial changes that mimic different writing habits.
For sentence-level perturbations, eight strategies are included to address more complex linguistic variations: Formal, Casual, Passive, and Active transformations from \cite{Li2018DeleteRG,Etinger2019FormalityST,Schmidt2020GenerativeTS,ChantrySchiappa2022MultimodalRA} alter the style of the text; Back Translation from \cite{ma2019nlpaug}; SCPN from \cite{iyyer2018adversarial}; BART from \cite{Lewis2019BARTDS}; and DIPPER from \cite{Krishna2023ParaphrasingED}, focus on semantic shifts due to translation errors and paraphrasing.
Similar to image perturbations, each text perturbation strategy is assigned severity levels. Character-level and word-level perturbations include five severity levels, mirroring the approach used for image perturbations, whereas sentence-level perturbations are applied at a single severity level. In total, this results in 63 text perturbation methods.
These techniques encompass a broad range of real-world text distribution shifts—such as typos, word swaps, and style changes
Detailed descriptions of each perturbation method are provided in Table~\ref{table:text_perturbation}  in Appendix~\ref{sec:appendix-perturbations}.

\section{Watermarks}
In this study, we utilize 4 image watermarking methods and 4 text watermarking methods to evaluate their robustness. The subsequent sections provide a brief introduction to each of these methods.

\vspace{-5pt}
\paragraph{Image Watermarks (Image-WM)}
\vspace{-5pt}
\begin{itemize}
    \item \textbf{DwtDctSvd-WM} \cite{cox2007digital}  integrates Discrete Wavelet Transform (DWT), Discrete Cosine Transform (DCT), and Singular Value Decomposition (SVD) to embed watermarks in color images. It starts by converting the RGB cover image to YUV, applies DWT to the Y channel, segments it into blocks via DCT, and performs SVD on each block before embedding the watermark. DwtDctSvd is the default method used by Stable Diffusion \cite{2022stablediffusion}.
    \vspace{-2pt}
    \item \textbf{RivaGAN-WM} \cite{Zhang2019RobustIV} introduces a robust image watermarking technique utilizing Generative Adversarial Networks (GANs) \cite{Goodfellow2014GenerativeAN}. It incorporates two adversarial networks: one to evaluate the quality of watermarked images and another to facilitate watermark removal. The system includes an encoder for watermark embedding and a decoder for its extraction, enhancing both performance and robustness. RivaGAN is also employed as a watermarking method by Stable Diffusion \cite{2022stablediffusion}.
    \vspace{-2pt}
    \item \textbf{SSL-WM} \cite{Fernandez2021WatermarkingII} leverages the latent spaces of pre-trained neural networks for watermark encoding, using networks trained via self-supervised learning (SSL) to capture effective watermarking features. This method embeds watermarks by applying backpropagation and data augmentation, and it is capable of detecting and decoding these watermarks from the watermarked image or its extracted features.
    \vspace{-2pt}
    \item \textbf{StegaStamp-WM} \cite{Tancik2019StegaStampIH} introduces a learned steganographic algorithm designed for the robust encoding and decoding of arbitrary hyperlink bit strings into photos, achieving near-perceptual invisibility. The system utilizes a deep neural network to learn an encoding/decoding algorithm that remains robust against image perturbations typical of real-world printing and photography scenarios.
\end{itemize}

\vspace{-5pt}
\paragraph{Text Watermarks (Text-WM)}
\begin{itemize}
    \item \textbf{KGW-WM} \cite{Kirchenbauer2023AWF} randomly splits the vocabulary into red tokens and green tokens based on the hash value of the previous token. During the next token generation, a constant $\delta$ is added to the logits for tokens that belong to the green list. This effectively increases the probability of generating green list tokens, thereby increasing the overall number of green tokens in the entire output sequence. During detection, if the suspect text contains significantly more green tokens, it is likely from the watermarked LLM.
    \vspace{-2pt}
    \item \textbf{KTH-WM} obtains a watermark key from the watermark sequence and generates the watermarked text by mapping the watermark key (random numbers) to the sample from the language model. \cite{Kuditipudi2023RobustDW} provided two instantiations of this protocol, namely inverse transform sampling and exponential minimum sampling. Both schemes ensure distortion-free -- the expected distribution of a single response from the watermarked model is identical to the distribution of a single response from the original model. We utilize exponential minimum sampling for our experiments due to its stronger reported results \cite{Kuditipudi2023RobustDW}.
    \vspace{-2pt}
    \item \textbf{Blackbox-WM} \cite{Yang2023WatermarkingTG}: begins by obtaining original text from a black-box language model. The process involves selectively replacing words with context-based synonyms to embed a watermark. It employs a unique binary encoding function that assigns a random binary value (either bit-0 or bit-1) to each word. This function ensures a near balance between bit-0 and bit-1 representations in non-watermarked texts. For every word selected for replacement, the method generates synonym candidates, each evaluated for the binary encoding they carry.
    \vspace{-2pt}
    \item \textbf{Unigram-WM} \cite{zhao2023provable} involves a watermarking process similar to KGW-WM, splitting the vocabulary into the green list and the red list and then increasing the probability of generating green tokens. The key difference is that Unigram-WM keeps the red-green partitions fixed. This allows for better robustness guarantees since each edit (insertion/deletion/replacement) only changes one token from green to red or from red to green.
\end{itemize}

\begin{table}[tp]\centering
\caption{Comparison of different image watermarking methods.}
\begin{adjustbox}{width=0.9\linewidth}
\begin{tabular}{llccccccc}\toprule
\textbf{Watermark} &\textbf{Model} &\textbf{PSNR} &\textbf{SSIM} &\textbf{Bit Acc} &\textbf{Dect Acc} &\textbf{Dect Acc (Ori)} &\cellcolor[HTML]{fff2cc}\textbf{Dect Acc Drop (\%)} \\
\midrule
\multirow{8}{*}{DctDwtSvd-WM} &NextGPT &17.03 &0.50 &4.29 &12.12 &95.56 &\cellcolor[HTML]{fff2cc}-87.32\% \\
&Stable Diffusion &17.69 &0.51 &4.43 &15.79 &100.00 &\cellcolor[HTML]{fff2cc}-84.21\% \\
&DALLE3 &16.85 &0.51 &3.09 &15.81 &95.32 &\cellcolor[HTML]{fff2cc}-83.41\% \\
&SDXL-Lightning &18.55 &0.56 &5.96 &21.27 &100.00 &\cellcolor[HTML]{fff2cc}-78.73\% \\
&PIXART &17.66 &0.54 &4.28 &18.83 &95.57 &\cellcolor[HTML]{fff2cc}-80.30\% \\
&Kandinsky 2.2 &15.99 &0.50 &4.01 &7.51 &91.42 &\cellcolor[HTML]{fff2cc}-91.79\% \\
&LCMs &19.48 &0.58 &6.32 &18.81 &100.00 &\cellcolor[HTML]{fff2cc}-81.19\% \\
&RPG-Image &17.74 &0.50 &3.79 &29.19 &92.20 &\cellcolor[HTML]{fff2cc}-68.34\% \\
\midrule
\multirow{8}{*}{RivaGAN-WM} &NextGPT &17.60 &0.54 &3.87 &23.40 &98.55 &\cellcolor[HTML]{fff2cc}-76.26\% \\
&Stable Diffusion &17.19 &0.51 &3.10 &31.39 &100.00 &\cellcolor[HTML]{fff2cc}-68.61\% \\
&DALLE3 &16.93 &0.52 &3.59 &25.40 &97.59 &\cellcolor[HTML]{fff2cc}-73.97\% \\
&SDXL-Lightning &18.40 &0.55 &3.56 &36.83 &100.00 &\cellcolor[HTML]{fff2cc}-63.17\% \\
&PIXART &17.82 &0.55 &7.01 &33.10 &96.63 &\cellcolor[HTML]{fff2cc}-65.75\% \\
&Kandinsky 2.2 &16.05 &0.51 &5.53 &17.17 &90.33 &\cellcolor[HTML]{fff2cc}-80.99\% \\
&LCMs &19.28 &0.57 &3.83 &35.84 &99.00 &\cellcolor[HTML]{fff2cc}-63.80\% \\
&RPG-Image &17.30 &0.50 &4.29 &29.55 &91.63 &\cellcolor[HTML]{fff2cc}-67.75\% \\
\midrule
\multirow{8}{*}{SSL-WM} &NextGPT &7.71 &0.30 &7.05 &26.18 &96.67 &\cellcolor[HTML]{fff2cc}-72.92\% \\
&Stable Diffusion &8.08 &0.31 &2.51 &24.72 &95.42 &\cellcolor[HTML]{fff2cc}-74.09\% \\
&DALLE3 &9.91 &0.27 &3.62 &25.67 &88.54 &\cellcolor[HTML]{fff2cc}-71.01\% \\
&SDXL-Lightning &10.65 &0.32 &3.62 &28.14 &86.48 &\cellcolor[HTML]{fff2cc}-67.46\% \\
&PIXART &8.54 &0.30 &3.09 &24.86 &92.41 &\cellcolor[HTML]{fff2cc}-73.10\% \\
&Kandinsky 2.2 &8.46 &0.24 &3.90 &20.91 &90.85 &\cellcolor[HTML]{fff2cc}-76.98\% \\
&LCMs &8.75 &0.34 &3.06 &12.01 &94.87 &\cellcolor[HTML]{fff2cc}-87.34\% \\
&RPG-Image &10.54 &0.30 &3.40 &23.70 &90.18 &\cellcolor[HTML]{fff2cc}-73.72\% \\
\midrule
\multirow{8}{*}{StegaStamp-WM} &NextGPT &7.33 &0.24 &3.49 &14.12 &85.73 &\cellcolor[HTML]{fff2cc}-83.53\% \\
&Stable Diffusion &7.38 &0.29 &4.04 &19.26 &82.29 &\cellcolor[HTML]{fff2cc}-76.59\% \\
&DALLE3 &6.83 &0.29 &3.84 &15.98 &88.41 &\cellcolor[HTML]{fff2cc}-81.93\% \\
&SDXL-Lightning &6.6 &0.22 &4.76 &9.25 &84.72 &\cellcolor[HTML]{fff2cc}-89.08\% \\
&PIXART &8.15 &0.21 &3.31 &14.53 &85.69 &\cellcolor[HTML]{fff2cc}-83.04\% \\
&Kandinsky 2.2 &6.91 &0.22 &3.68 &9.64 &86.35 &\cellcolor[HTML]{fff2cc}-88.84\% \\
&LCMs &6.94 &0.21 &4.15 &25.71 &88.67 &\cellcolor[HTML]{fff2cc}-85.78\% \\
&RPG-Image &5.81 &0.26 &3.46 &8.74 &80.26 &\cellcolor[HTML]{fff2cc}-89.11\% \\
\bottomrule
\end{tabular}
\end{adjustbox}
\label{table:image_watermark_results}
\end{table}

\section{Experiments}

In this study, we aim to address several key questions: (1) Which watermarking methods demonstrate the highest stability and robustness? (2) Which perturbation techniques are most effective? (3) Among image and text generation models, which are the most robust? 

\vspace{-5pt}
\subsection{Experimental Settings}\label{sec:exp_setting}

\vspace{-5pt}
\paragraph{Benchmark Models}

We investigate the following benchmark models, all of which are publicly available. Each model has been chosen for its relevance and potential to provide insights into the watermark robustness against perturbations. We used 16 NVIDIA A6000 GPUs for our experiments.
\begin{itemize}[leftmargin=*, itemsep=0pt, topsep=0pt]
\item Text-to-image models: 
NExT-GPT \cite{wu2023next}, 
Stable Diffusion \cite{Rombach2021HighResolutionIS}, 
DALLE3 \cite{BetkerImprovingIG}, 
SDXL-Lightning \cite{Lin2024SDXLLightningPA}, 
PIXART \cite{Chen2023PixArtFT}, 
Kandinsky 2.2 \cite{Razzhigaev2023KandinskyAI}, 
Latent Consistency Models (LCMs) \cite{Luo2023LatentCM},
RPG \cite{Yang2024MasteringTD}.
\vspace{-2pt}
\item Image-to-text models: 
NExT-GPT \cite{wu2023next},
Fuyu-8B \cite{fuyu-8b},
InternLM-XComposer \cite{Zhang2023InternLMXComposerAV},
InstructBLIP \cite{Dai2023InstructBLIPTG},
LLaVA 1.6 \cite{Liu2023ImprovedBW}, 
MiniGPT-4 \cite{Zhu2023MiniGPT4EV},
mPLUG-Owl2 \cite{Ye2023mPLUGOwl2RM},
Qwen-VL \cite{Bai2023QwenVLAF}.
\end{itemize}

\vspace{-5pt}
\paragraph{Evaluation Metrics}

For evaluating image quality, our study employs several metrics: Peak Signal-to-Noise Ratio (PSNR), Structural Similarity Index Measure (SSIM) \cite{Wang2004ImageQA}, Bit Accuracy (Bit Acc), and Detection Accuracy (Dect Acc) \cite{zhao2023provable}. In assessing text quality, we utilize BLEURT \cite{Sellam2020BLEURTLR,Pu2021LearningCM}, ROUGE \cite{Lin2004ROUGEAP}, Bit Accuracy (Bit Acc), and Detection Accuracy (Dect Acc) \cite{zhao2023provable}. These metrics allow us to assess the fidelity and integrity of watermarked images comprehensively.

\begin{table}[tp]\centering
\caption{Comparison of different text watermarking methods.}
\begin{adjustbox}{width=0.95\linewidth}
\begin{tabular}{llccccccc}\toprule
\textbf{Watermark} &\textbf{Model} &\textbf{BLEURT} &\textbf{ROUGE} &\textbf{Bit Acc} &\textbf{Dect Acc} &\textbf{Dect Acc (ori)} &\cellcolor[HTML]{fff2cc}\textbf{Dect Acc Drop (\%)} \\
\midrule
\multirow{8}{*}{KGW-WM} &NextGPT &0.32 &39.54 &59.41 &19.60 &99.76 &\cellcolor[HTML]{fff2cc}-80.35\% \\
&Fuyu &0.30 &40.39 &67.87 &36.27 &99.85 &\cellcolor[HTML]{fff2cc}-63.68\% \\
&InternLM-XComposer &0.31 &35.07 &64.68 &14.97 &100.00 &\cellcolor[HTML]{fff2cc}-85.03\% \\
&InstructBLIP &0.30 &3.07 &51.26 &23.60 &100.00 &\cellcolor[HTML]{fff2cc}-76.40\% \\
&LLaVA 1.5 &0.35 &42.34 &68.11 &71.38 &98.04 &\cellcolor[HTML]{fff2cc}-37.93\% \\
&MiniGPT-4 &0.37 &43.21 &57.84 &18.95 &99.48 &\cellcolor[HTML]{fff2cc}-80.95\% \\
&mPLUG-Owl2 &0.30 &43.17 &59.78 &37.64 &99.95 &\cellcolor[HTML]{fff2cc}-62.34\% \\
&Qwen-VL &0.29 &29.16 &63.09 &23.10 &99.93 &\cellcolor[HTML]{fff2cc}-76.88\% \\
\midrule
\multirow{8}{*}{KTH-WM} &NextGPT &0.31 &41.43 &51.78 &21.06 &98.37 &\cellcolor[HTML]{fff2cc}-78.59\% \\
&Fuyu &0.29 &41.50 &66.34 &33.50 &99.06 &\cellcolor[HTML]{fff2cc}-66.18\% \\
&InternLM-XComposer &0.23 &38.13 &57.86 &33.00 &100.00 &\cellcolor[HTML]{fff2cc}-67.00\% \\
&InstructBLIP &0.36 &14.54 &56.34 &34.00 &99.85 &\cellcolor[HTML]{fff2cc}-65.95\% \\
&LLaVA 1.5 &0.31 &31.74 &64.90 &35.75 &98.68 &\cellcolor[HTML]{fff2cc}-63.77\% \\
&MiniGPT-4 &0.32 &44.14 &50.63 &21.50 &100.00 &\cellcolor[HTML]{fff2cc}-78.50\% \\
&mPLUG-Owl2 &0.27 &36.63 &62.55 &33.41 &99.64 &\cellcolor[HTML]{fff2cc}-66.47\% \\
&Qwen-VL &0.25 &35.52 &54.23 &32.59 &99.87 &\cellcolor[HTML]{fff2cc}-67.37\% \\
\midrule
\multirow{8}{*}{Blackbox-WM} &NextGPT &0.33 &43.09 &62.41 &23.80 &100.00 &\cellcolor[HTML]{fff2cc}-76.20\% \\
&Fuyu &0.30 &40.01 &66.12 &27.93 &100.00 &\cellcolor[HTML]{fff2cc}-72.07\% \\
&InternLM-XComposer &0.28 &33.44 &52.75 &25.40 &99.57 &\cellcolor[HTML]{fff2cc}-74.49\% \\
&InstructBLIP &0.37 &35.51 &53.24 &22.70 &99.86 &\cellcolor[HTML]{fff2cc}-77.27\% \\
&LLaVA 1.5 &0.31 &42.22 &60.60 &22.72 &99.62 &\cellcolor[HTML]{fff2cc}-77.19\% \\
&MiniGPT-4 &0.37 &44.72 &64.53 &24.01 &100.00 &\cellcolor[HTML]{fff2cc}-75.99\% \\
&mPLUG-Owl2 &0.30 &34.81 &62.54 &31.58 &99.86 &\cellcolor[HTML]{fff2cc}-68.38\% \\
&Qwen-VL &0.28 &33.19 &51.46 &24.68 &99.68 &\cellcolor[HTML]{fff2cc}-75.24\% \\
\midrule
\multirow{8}{*}{Unigram-WM} &NextGPT &0.26 &26.39 &48.42 &2.46 &100.00 &\cellcolor[HTML]{fff2cc}-97.54\% \\
&Fuyu &0.24 &28.14 &48.98 &0.77 &100.00 &\cellcolor[HTML]{fff2cc}-99.23\% \\
&InternLM-XComposer &0.19 &25.37 &47.47 &0.32 &100.00 &\cellcolor[HTML]{fff2cc}-99.68\% \\
&InstructBLIP &0.25 &17.18 &47.88 &17.74 &99.76 &\cellcolor[HTML]{fff2cc}-82.22\% \\
&LLaVA 1.5 &0.37 &40.01 &58.16 &4.68 &99.69 &\cellcolor[HTML]{fff2cc}-95.31\% \\
&MiniGPT-4 &0.32 &32.67 &45.33 &2.48 &98.62 &\cellcolor[HTML]{fff2cc}-97.49\% \\
&mPLUG-Owl2 &0.23 &28.08 &47.58 &1.27 &97.42 &\cellcolor[HTML]{fff2cc}-98.70\% \\
&Qwen-VL &0.21 &24.99 &42.57 &0.28 &99.03 &\cellcolor[HTML]{fff2cc}-99.72\% \\
\bottomrule
\end{tabular}
\end{adjustbox}
\label{table:text_watermark_results}
\end{table}

\subsection{Results And Discussions}

\begin{figure}[tp]
\centering
\begin{minipage}[t]{0.48\textwidth}
\centering
\includegraphics[width=0.99\linewidth]{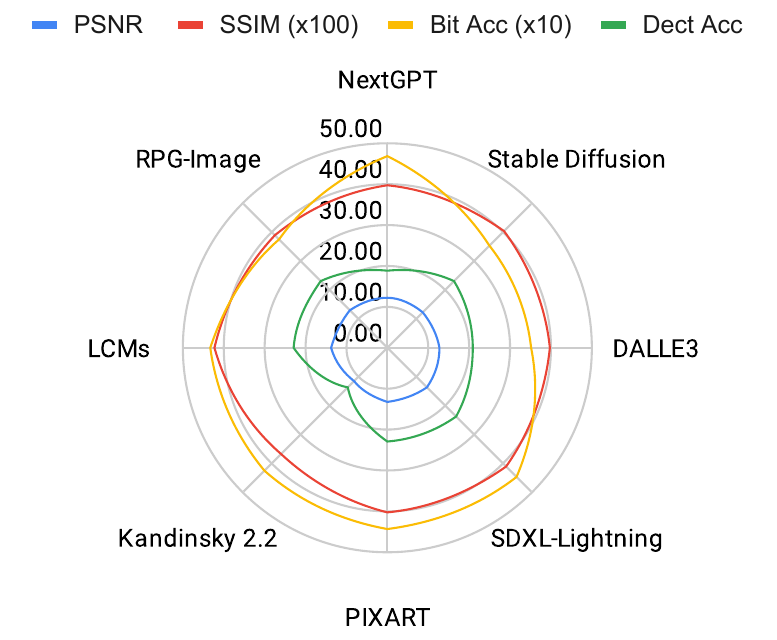}
\vspace{-5pt}
\caption{Performance comparison of different models under image perturbations.}
\label{Fig:image_radar}
\end{minipage}
~~
\begin{minipage}[t]{0.48\textwidth}
\centering
 \includegraphics[width=0.99\linewidth]{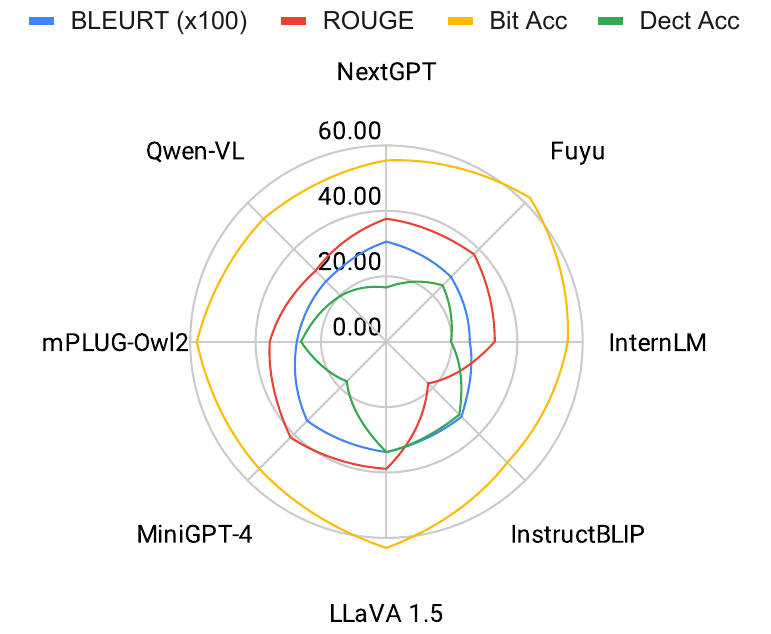}
 \vspace{-5pt}
  \caption{Performance comparison of different models under text perturbations.}
  \label{Fig:text_radar}
\end{minipage}
\vspace{-15pt}
\end{figure}

\begin{figure}[tp]
  \centering
  \includegraphics[width=0.99\linewidth]{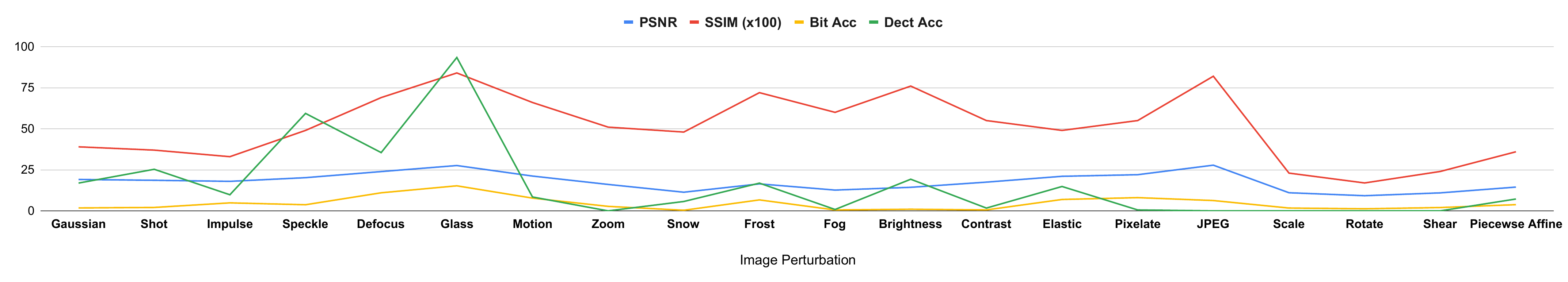}
  \includegraphics[width=0.99\linewidth]{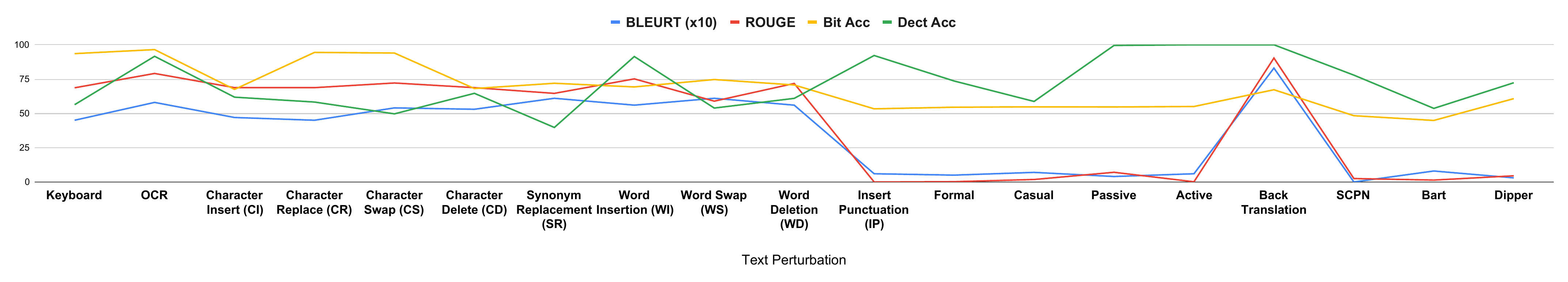}
  \vspace{-5pt}
  \caption{Comparisons of different [Top] image corruption and  [Bottom] text perturbation methods using Stable Diffusion and LLaVA, respectively. All the results have been averaged on different severity levels.}
  \label{Fig:comp_perturb}
  \vspace{-5pt}
\end{figure}

\vspace{-5pt}
\paragraph{Watermarking Strategy Comparison.}

In Tables~\ref{table:image_watermark_results} and \ref{table:text_watermark_results}, we present the outcomes of various image and text watermarking methods, respectively. The results indicate variations in the performance across different models. Generally, under image perturbations, RivaGAN-WM appears to be more robust, whereas under text perturbations, KGW-WM demonstrates greater stability.

\vspace{-5pt}
\paragraph{Perturbation Method Comparison.}

In Figure~\ref{Fig:comp_perturb}, we showed the comparisons of different [Top] image corruption and  [Bottom] text perturbation methods. All the results have been averaged on different severity levels. 

For image corruptions, we show the results by Stable Diffusion in Figure~\ref{Fig:comp_perturb} [Top]. 
Generally, noise and blur-based perturbations tend to have a more severe impact on all metrics, as they directly affect the clarity and sharpness of images. Environmental effects and quality degradation also impact the metrics but might be less severe compared to direct noise introductions or blurs. 
Gaussian, Shot, Impulse, and Speckle noise appear to significantly decrease PSNR and SSIM, indicating a substantial degradation in image quality. JPEG compression, while degrading quality, may not affect structural similarity as much, depending on the compression level.
Distortions such as Rotate, Shear, and Piecewise Affine could particularly lower Detection Accuracy as they alter the geometry and spatial relationships within the image, potentially complicating detection tasks.
Based on the results in Figure~\ref{Fig:comp_perturb} [Top], we find that Zoom Blur is more effective, and Glass Blur is less effective.

Similarly, for text perturbations, we show the results by LLaVA 1.6 in Figure~\ref{Fig:comp_perturb} [Bottom]. 
Stylistic transformations (Formal, Casual, Active, Passive) show the highest performance across all metrics, indicating minimal impact on text quality and excellent preservation and detection capabilities. Synonym Replacement (SR), Bart, SPCN, and Dipper also perform well, with high BLEURT and ROUGE scores and good preservation and detection accuracy. Character Insert (CI), Character Delete (CD), and Word Swap (WS) show the lowest scores across all metrics, indicating significant degradation in text quality and moderate preservation and detection capabilities.
Based on the results in Figure~\ref{Fig:comp_perturb} [Bottom], we find that character-level perturbations are more effective, and sentence-level perturbations are less effective.

\begin{figure}[tp]
  \centering
  \includegraphics[width=0.95\linewidth]{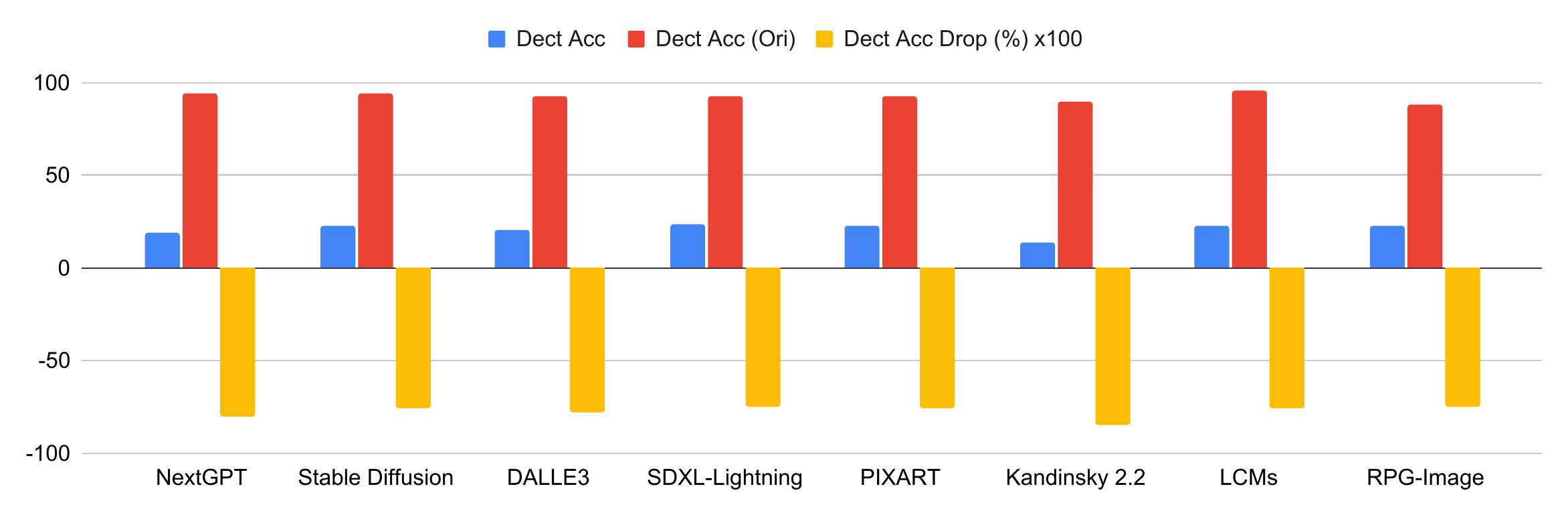}
  \includegraphics[width=0.95\linewidth]{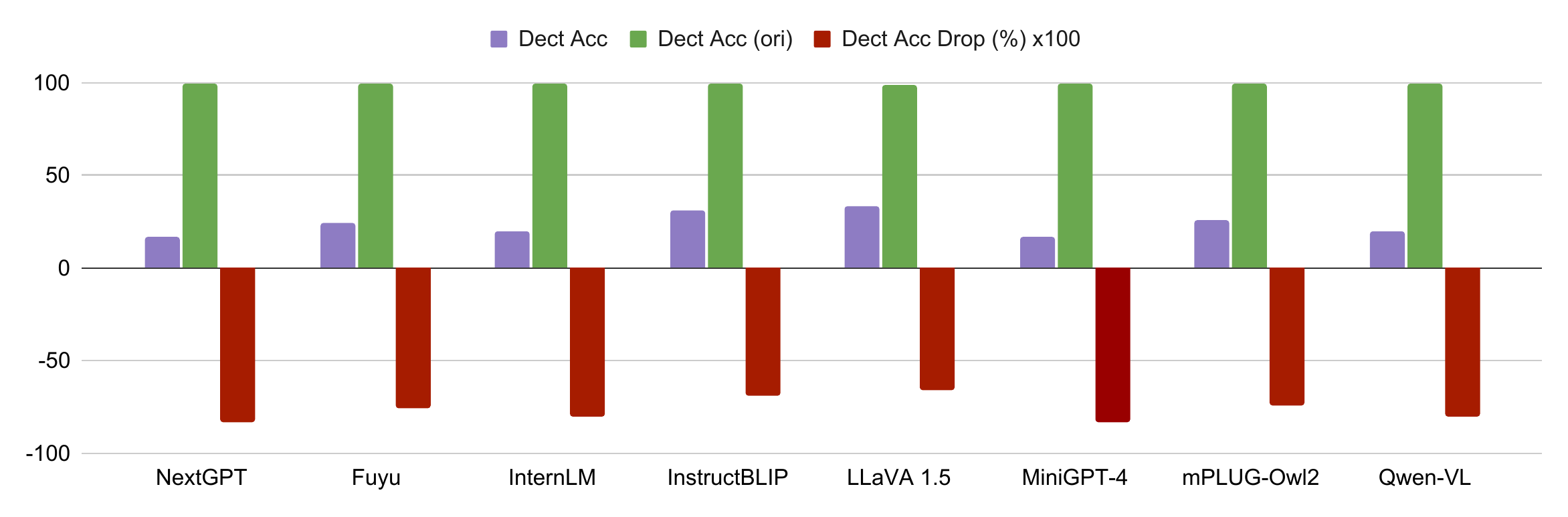}
  \vspace{-5pt}
  \caption{Model comparisons under [Top] image corrections and [Bottom] text perturbations. All the results have been averaged on the performance under all image/text perturbations. }
  \label{Fig:comp_model}
  \vspace{-15pt}
\end{figure}

\vspace{-5pt}
\paragraph{Model Comparison.}

In Figure~\ref{Fig:comp_model}, we showed the model comparisons under [Top] image and [Bottom] text perturbations. The results are averaged across all image and text perturbations. Our findings indicate that under image perturbations, SDXL-Lightning demonstrates superior robustness. Conversely, LLaVA exhibits greater robustness under text perturbations compared to the other models.

\vspace{-5pt}
\subsection{Ablation Study}

\paragraph{Image Perturbation Severity Influence.}

\begin{figure}[tp]
  \centering
  \includegraphics[width=0.9\linewidth]{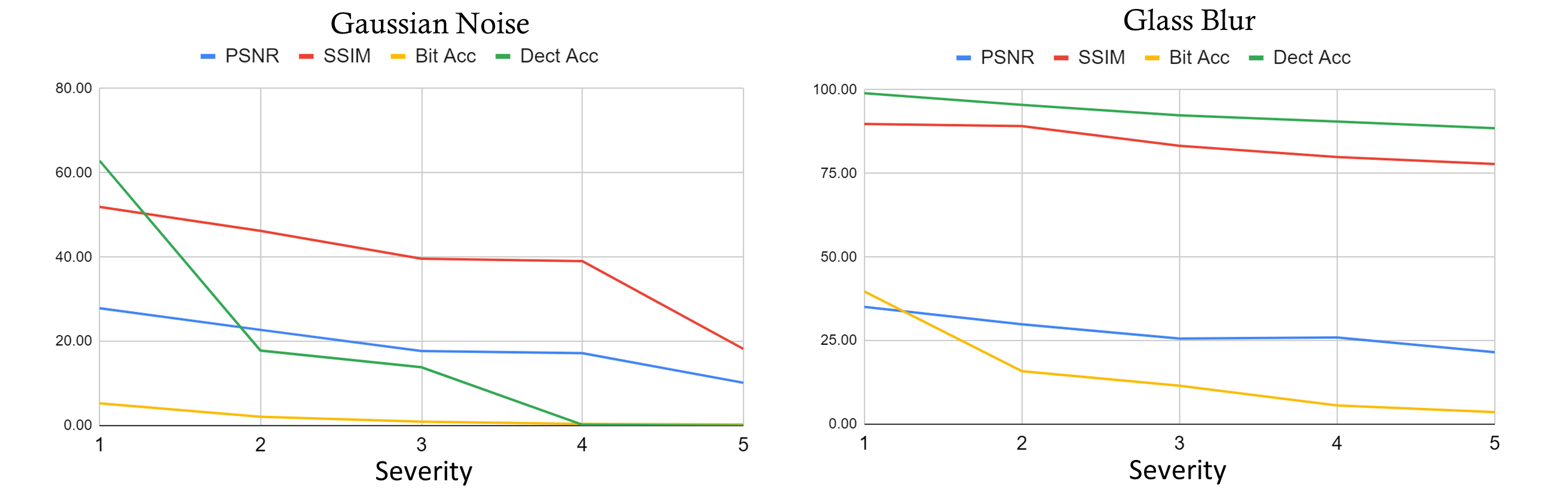}
  \caption{Performance changes with different severity levels under image perturbations.}
  \label{Fig:perturb_severity_image}
  \vspace{-5pt}
\end{figure}

Each image perturbation method is associated with multiple severity levels, so we would like to explore the relationship between robustness performance and perturbation severity levels. In Figure~\ref{Fig:perturb_severity_image}, we showed two examples, Gaussian Noise and Glass Blur, across varying levels of severity (from 1 to 5). PSNR and SSIM are particularly sensitive to distortions, with PSNR showing a more pronounced drop in the presence of Gaussian Noise than Glass Blur. This suggests that noise introduces more disruptive interference compared to blur. SSIM, while also decreasing with severity, indicates that structural elements of images are somewhat more preserved under Glass Blur compared to Gaussian Noise, highlighting differential impacts depending on the type of distortion.
Bit Accuracy exhibits a notable decline under both conditions but is more affected by Glass Blur, especially beyond moderate levels of severity. This suggests that blur more significantly affects the bit-level representation of the image. Detection Accuracy remains relatively stable under Glass Blur and shows resilience, suggesting that detection algorithms might still identify key features in blurred images effectively. However, it declines under Gaussian Noise, indicating challenges in feature detection amidst this type of noise.

\vspace{-5pt}
\paragraph{Text Perturbation Severity Influence.}

\begin{figure}[tp]
  \centering
  \includegraphics[width=0.9\linewidth]{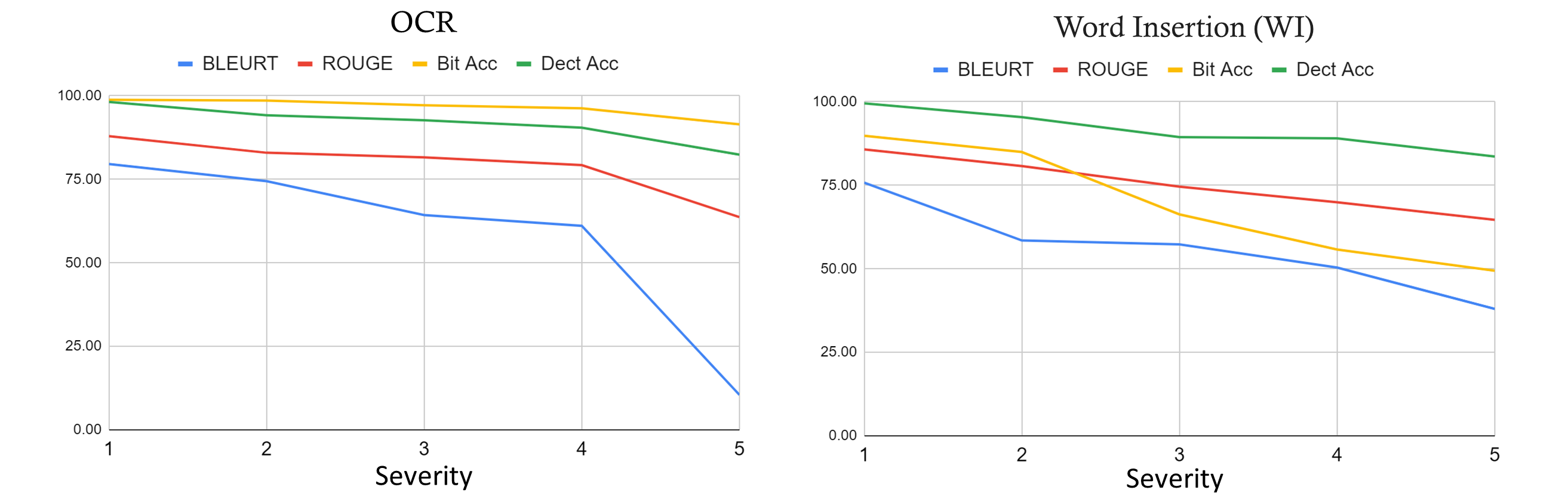}
  \caption{Performance changes with different severity levels under text perturbations.}
  \label{Fig:perturb_severity_text}
  \vspace{-10pt}
\end{figure}

Similar to the image perturbations above, in Figure~\ref{Fig:perturb_severity_text}, we showed two examples, OCR and Word Insertion, across varying levels of severity (from 1 to 5).  OCR errors cause a more drastic decrease in BLEURT scores than word insertions, suggesting that OCR errors might lead to more severe semantic disruptions than simple word additions. Detection accuracy demonstrates notable resilience in both scenarios, implying that the underlying algorithms are effective at extracting essential information despite textual distortions.

\vspace{-5pt}
\subsection{Limitations and Future Work.}

\begin{itemize}[leftmargin=*, itemsep=0pt, topsep=0pt]
    \item Our current evaluation study focuses on watermark robustness against image corruptions and text perturbations. However, it does not encompass other watermark types, such as audio, nor does it include all potential perturbations. For instance, while there are studies on the robustness against adversarial image perturbations, such attacks typically involve a classification task conditioned on a target label, making them inapplicable to our evaluation setting. Therefore, they were not included in this study.
    \item Our study investigated 8 image-to-text models and 8 text-to-image models. However, we acknowledge that this is a preliminary study. Different watermarking strategies may show varying performance when the embedded watermarks change. Consequently, the findings related to the models are speculative and not conclusive.
    \item We hope that future research will expand the evaluation benchmark to include a broader range of watermark types and perturbations. To the best of our knowledge, our study does not pose any potential negative societal impacts.
\end{itemize}


\takeaway{Our main findings are as follows. \\
(1) Multimodal watermarks are sensitive to image corruptions and text perturbations. \\
(2) Among image perturbations, Zoom Blur consistently shows the highest impact, while Glass Blur is the least harmful. \\
(3) Among text perturbations, character-level perturbations are more effective, and sentence-level perturbations are less effective. \\
(4) Under image perturbations, RivaGAN-WM appears more stable, whereas under text perturbations, KGW-WM seems more stable. \\
(5) In terms of models, SDXL-Lightning is more robust than other baselines under image perturbations, while LLaVA demonstrates greater robustness under text perturbations.
}

\vspace{-5pt}
\section{Conclusion}

In this study, we explore the robustness of multimodal watermarks to perturbations in images and text. Our research includes testing the performance of 8 image-to-text models and 8 text-to-image models, subjected to 100 image perturbation techniques and 63 text perturbation methods. We assess the robustness of 4 image watermarking methods and 4 text watermarking methods. We aim for our benchmark to be a valuable resource for examining the robustness of multimodal watermarks, and we hope our results will inspire the development and implementation of more robust multimodal watermarking strategies for practical applications.

\clearpage
\section*{Acknowlegement}
    This study is supported in part by Google Research and Carnegie Mellon University Computer Science Department fellowship. We would like to thank the feedback from Sven Gowal, Yonatan Bisk, Daniel Fried, and William Wang.

\bibliography{custom}

\clearpage
\appendix
\onecolumn

\section{More Details about Perturbation Methods}\label{sec:appendix-perturbations}


Table~\ref{table:image_perturbation} shows a detailed introduction to each image perturbation method.   Table~\ref{table:text_perturbation} shows a detailed introduction about each text perturbation method. 

\vspace{-5pt}
\begin{table*}[htp]\small
\centering
\caption{Image perturbations.}
\vspace{5pt}
    \begin{adjustbox}{width=0.99\linewidth}
    \begin{tabular}{ll|p{8cm}|c}
    \toprule
      Category & Perturbation  & Description &Severities  \\ 
    \midrule
     \multirow{7}{*}{Noise } 
     &Gaussian Noise  &Gaussian noise can appear in low-lighting conditions. 	&5	 \\
     \cmidrule(l){2-2} \cmidrule(l){3-3} \cmidrule(l){4-4}
     &Shot Noise  &Shot noise, also called Poisson noise, is electronic noise caused by the discrete nature of light itself. 	&5	 \\ 
     \cmidrule(l){2-2} \cmidrule(l){3-3} \cmidrule(l){4-4}
     &Impulse Noise  &Impulse noise is a color analogue of salt-and-pepper noise and can be caused by bit errors. 	&5	 \\ 
     \cmidrule(l){2-2} \cmidrule(l){3-3} \cmidrule(l){4-4}
     &Speckle Noise  &Speckle noise is the noise added to a pixel that tends to be larger if the original pixel intensity is larger.	&5	 \\
     \midrule
     \multirow{6}{*}{Blur}
     &Defocus Blur  &Defocus blur occurs when an image is out of focus. 	&5	 \\  
     \cmidrule(l){2-2} \cmidrule(l){3-3} \cmidrule(l){4-4}
     &Frosted Glass Blur  &Frosted Glass Blur appears with “frosted glass” windows or panels. 	&5	 \\ 
     \cmidrule(l){2-2} \cmidrule(l){3-3} \cmidrule(l){4-4}
     &Motion Blur  &Motion blur appears when a camera is moving quickly. 	&5	 \\ 
     \cmidrule(l){2-2} \cmidrule(l){3-3} \cmidrule(l){4-4}
     &Zoom Blur  &Zoom blur occurs when a camera moves toward an object rapidly.  	&5	 \\ 
     \midrule
     \multirow{6}{*}{Weather} 
     &Snow  &Snow is a visually obstructive form of precipitation.	&5	  \\
     \cmidrule(l){2-2} \cmidrule(l){3-3} \cmidrule(l){4-4}
     &Frost  &Frost forms when lenses or windows are coated with ice crystals. 	&5	  \\
     \cmidrule(l){2-2} \cmidrule(l){3-3} \cmidrule(l){4-4}
     &Fog  &Fog shrouds objects and is rendered with the diamond-square algorithm.	&5	  \\
     \cmidrule(l){2-2} \cmidrule(l){3-3} \cmidrule(l){4-4}
     &Brightness  &Brightness varies with daylight intensity. 	&5	  \\
     \midrule
     \multirow{6}{*}{Digital} 
     &Contrast  &Contrast can be high or low depending on lighting conditions and the photographed object’s color. 	&5	  \\
     \cmidrule(l){2-2} \cmidrule(l){3-3} \cmidrule(l){4-4}
     &Elastic  &Elastic transformations stretch or contract small image regions. 	&5	  \\
     \cmidrule(l){2-2} \cmidrule(l){3-3} \cmidrule(l){4-4}
     &Pixelate  &Pixelation occurs when upsampling a low-resolution image.	&5	  \\
     \cmidrule(l){2-2} \cmidrule(l){3-3} \cmidrule(l){4-4}
     &JPEG Compression  &JPEG is a lossy image compression format that introduces compression artifacts.	&5	  \\
     \midrule
     \multirow{9}{*}{Geometric}
     &Scale &Change the size of an image by enlarging or shrinking its dimensions.   &5 \\
      \cmidrule(l){2-2} \cmidrule(l){3-3} \cmidrule(l){4-4}
     &Rotate &Turn the image around a central point by a specified degree, altering its orientation.   &5 \\
     \cmidrule(l){2-2} \cmidrule(l){3-3} \cmidrule(l){4-4}
      & Shear &Skew the image by shifting parts of it more than others, creating a distortion.   &5 \\
      \cmidrule(l){2-2} \cmidrule(l){3-3} \cmidrule(l){4-4}
      & Piecewise Affine &Apply affine transformations to different parts of the image independently, allowing for complex local distortions.  &5 \\
    \midrule
    Sum & \textbf{20} & --- &\textbf{100} \\
    \bottomrule
\end{tabular}
    \end{adjustbox}
\label{table:image_perturbation}
\vspace{-5pt}
\end{table*}



\begin{table*}[tp]\small
\centering
\caption{Text perturbations.}
\vspace{5pt}
    \begin{adjustbox}{width=0.999\linewidth}
    \begin{tabular}{ll|p{7cm}|c}
    \toprule
      Category & Perturbation  & Description &Severities  \\ 
    \midrule
     \multirow{9}{*}{Character-level } 
     &Keyboard  &Substitute character by keyboard distance with probability $p$.	&5	 \\
     \cmidrule(l){2-2} \cmidrule(l){3-3} \cmidrule(l){4-4}
     &OCR  &Substitute character by pre-defined OCR error with probability $p$.	&5	 \\ 
     \cmidrule(l){2-2} \cmidrule(l){3-3} \cmidrule(l){4-4}
     &Character Insert (CI)  &Insert character randomly with probability $p$.	&5	 \\
     \cmidrule(l){2-2} \cmidrule(l){3-3} \cmidrule(l){4-4}
     &Character Replace (CR)  &Substitute character randomly with probability $p$.	&5	 \\
     \cmidrule(l){2-2} \cmidrule(l){3-3} \cmidrule(l){4-4}
     &Character Swap (CS) &Swap character randomly with probability $p$.	&5	 \\ 
     \cmidrule(l){2-2} \cmidrule(l){3-3} \cmidrule(l){4-4}
     &Character Delete (CD) &Delete character randomly with probability $p$.	&5	 \\
     \midrule
     \multirow{14}{*}{Word-level}
     &Synonym Replacement (SR)  &Randomly choose $n$ words from the sentence that are not stop words. Replace each of these words with one of its synonyms chosen at random.	&5	 \\  
     \cmidrule(l){2-2} \cmidrule(l){3-3} \cmidrule(l){4-4}
     &Word Insertion (WI)  &Find a random synonym of a random word in the sentence that is not a stop word. Insert that synonym into a random position in the sentence. Do this $n$ times.	&5	 \\ 
     \cmidrule(l){2-2} \cmidrule(l){3-3} \cmidrule(l){4-4}
     &Word Swap (WS)  &Randomly choose two words in the sentence and swap their positions. Do this $n$ times.	&5	 \\ 
     \cmidrule(l){2-2} \cmidrule(l){3-3} \cmidrule(l){4-4}
     &Word Deletion (WD)  &Each word in the sentence can be randomly removed with probability $p$.	&5	 \\ 
     \cmidrule(l){2-2} \cmidrule(l){3-3} \cmidrule(l){4-4}
     &Insert Punctuation (IP)  &Random insert punctuation in the sentence with probability $p$.	&5	 \\ 
     \midrule
     \multirow{17}{*}{Sentence-level } 
     &Formal  &Transfer the text style to Formal.	&1	  \\
     \cmidrule(l){2-2} \cmidrule(l){3-3} \cmidrule(l){4-4}
     &Casual  &Transfer the text style to Casual.	&1	  \\
     \cmidrule(l){2-2} \cmidrule(l){3-3} \cmidrule(l){4-4}
     &Passive  &Transfer the text style to Passive.	&1	  \\
     \cmidrule(l){2-2} \cmidrule(l){3-3} \cmidrule(l){4-4}
     &Active  &Transfer the text style to Active.	&1	  \\
     \cmidrule(l){2-2} \cmidrule(l){3-3} \cmidrule(l){4-4}
     &Back Translation (BT) &Translate source to German and translate it back to English via \cite{wmt19-en-de}. &1	  \\
     \cmidrule(l){2-2} \cmidrule(l){3-3} \cmidrule(l){4-4}
     & SCPN  &Produce a paraphrase of a given sentence with specified syntactic structures \cite{iyyer2018adversarial}. &1 \\
     \cmidrule(l){2-2} \cmidrule(l){3-3} \cmidrule(l){4-4}
     & BART  &Use BART for text summarization as paraphrasing attack \cite{Lewis2019BARTDS}. &1\\
     \cmidrule(l){2-2} \cmidrule(l){3-3} \cmidrule(l){4-4}
     & DIPPER  &DIPPER can paraphrase paragraphs, condition on surrounding context, and control lexical diversity and content reordering \cite{Krishna2023ParaphrasingED}. &1\\
    \midrule
    Sum & \textbf{19} & --- &\textbf{63} \\
    \bottomrule
\end{tabular}
    \end{adjustbox}
\label{table:text_perturbation}
\end{table*}

\section{Experimental Settings}


\subsection{Image Watermark Parameters}

\vspace{-5pt}
\paragraph{DctDwtSvd-WM and RivaGAN-WM}
For all of our experiments, we embed 4 characters as bytes, specifically the string ``test", into the images. During decoding, we deem the watermark detection algorithm as passing if it is able to fully retrieve the original, encoded string ``test".

\vspace{-5pt}
\paragraph{SSL-WM}
A neural network is needed to extract the features from images, and a normalization layer is to evenly distribute the extracted features in the latent space.
We utilize the recommended, default model and normalization layers, namely ResNet-50 trained with DINO and PCA whitening respectively.
During detection, we consider the zero-bit scenario \cite{Fernandez2021WatermarkingII}.

\vspace{-5pt}
\paragraph{StegaStamp-WM}
requires a pretrained encoder, decoder, and detector.
The encoder is an architecture similar to U-Net \cite{ronneberger2015unet} where the image is first processed through a fully connected network to become a tensor of size $50 \times 50 \times 3$, then upsampled to get a tensor of size $400 \times 400 \times 3$. 
The decoder is a spatial transformer network \cite{NIPS2015_33ceb07b} that is trained to recover the encoded watermark. 
For the detector, the authors use an open-source semantic segmentation network, namely BiSeNet \cite{yu2018bisenet}.
We utilize all of the default parameters used in \cite{Tancik2019StegaStampIH}.

\vspace{-5pt}
\subsection{Text Watermark Parameters}

\vspace{-5pt}
\paragraph{KGW-WM}
We utilize game and delta values of 0.25 and 2.0, respectively.
We also set the seeding scheme as `simple\_1', which represents a simple bigram hash, to utilize the main settings of the experiments in the paper \cite{Kirchenbauer2023AWF}.
During detection, we also utilize a $z$ threshold of 0.5 and ignore repeated n-grams.

\vspace{-5pt}
\paragraph{KTH-WM}
We set the desired length of the generated text, $m$, to 30 as detailed in \cite{Kuditipudi2023RobustDW}.
The length of the watermark sequence, $n$, is kept at the standard value of 256. For generating the random watermark sequence, we employ a key of 42.
The authors' method of evaluating their watermarking framework involves p-values, and we consider texts with $p < 0.1$ as effectively watermarked.

\vspace{-5pt}
\paragraph{Blackbox-WM}
We employ the $\tau$ word value of 0.8, and a $\lambda$ value of 0.83 \cite{Yang2023WatermarkingTG}. 
We also use the ``embed" mode during watermarking.
During detection, if the confidence value is over 80\%, we deem the detection algorithm as successfully finding the watermark.

\vspace{-5pt}
\paragraph{Unigram-WM}
When applying the watermark, we utilize a fraction and strength value of 0.5 and 2.0, respectively. Additionally, we determined the watermark key to be defaulted to 0. During detection, we utilize the default value of 6.0 \cite{zhao2023provable}.

\subsection{Number of Samples}

We utilized 5,000 image-caption pairs from the COCO validation split \cite{lin2014microsoft}. For text generation, the input to the multimodal models consisted of the prompt ``Describe this image:" alongside the corresponding image from the dataset. For image generation, the input was the prompt ``Please generate an image describing the following caption: {$ \tt C$}", where $\tt C$ is the corresponding caption from the dataset. In total, we generated 5,000 images and 5,000 texts for each model.

\section{More Related Work}

\paragraph{Multimodal Learning}

Research in formalized multimodal learning dates back to 1989, initiated by an experiment by \cite{yuhas_1989_integration}, which expanded upon the McGurk Effect in the realm of audio-visual speech recognition using neural networks \citep{tiippana_2014_what, McGurk1976HearingLA}. Collaboration between experts in Natural Language Processing (NLP) and Computer Vision (CV) led to the creation of large, multimodal datasets designed for specific downstream tasks like classification, translation, and detection.
Subsequent enhancements in Large Language Models (LLMs) have paved the way for incorporating additional data modalities, most notably visual data \citep{wang_2022_unifying, qiu2023benchmarking, Nguyen2022GRITFA, li_2022_blip, wang_2021_simvlm, qiu2023sccs, Shah2022LMNavRN, Zhang_2021_CVPR, He2023AlignAA, qiu2024snapntell, qiu2024entity6k}. By leveraging embeddings that have been pretrained on both language and image datasets, vision-language models have demonstrated significant performance.

\vspace{-5pt}
\section{Detailed Experimental Results}

In the following tables, we provide the detailed results for each baseline model mentioned in Section~\ref{sec:exp_setting}.

\begin{table*}[htp]\small
\centering
\caption{NExT-GPT image watermarks under image perturbations.}
\begin{adjustbox}{width=0.60\linewidth}
\begin{tabular}{lllrrrrr}\toprule
\textbf{Watermark} &\textbf{Attack Category} &\textbf{Perturbation} &\textbf{PSNR} &\textbf{SSIM} &\textbf{Bit Acc} &\textbf{Dect Acc} \\
\midrule
\multirow{20}{*}{DctDwtSvd-WM} &\multirow{4}{*}{Noise} &Gaussian &18.95 &0.40 &1.87 &5.68 \\
& &Shot &18.32 &0.41 &2.42 &9.25 \\
& &Impulse &18.18 &0.38 &4.79 &1.38 \\
& &Speckle &0.00 &0.50 &3.59 &25.48 \\
\cmidrule(l){2-7}
&\multirow{4}{*}{Blur} &Defocus &22.93 &0.63 &7.40 &28.82 \\
& &Glass &27.05 &0.83 &10.56 &80.69 \\
& &Motion &20.53 &0.61 &5.92 &6.75 \\
& &Zoom &17.31 &0.50 &3.63 &0.00 \\
\cmidrule(l){2-7}
&\multirow{4}{*}{Weather} &Snow &11.13 &0.52 &0.29 &6.42 \\
& &Frost &10.00 &0.60 &0.33 &2.98 \\
& &Fog &13.22 &0.59 &0.70 &0.38 \\
& &Brightness &14.03 &0.77 &0.88 &18.75 \\
\cmidrule(l){2-7}
&\multirow{4}{*}{Digital} &Contrast &14.73 &0.53 &0.88 &1.48 \\
& &Elastic &16.70 &0.41 &5.02 &11.49 \\
& &Pixelate &22.76 &0.65 &7.97 &36.31 \\
& &JPEG &28.09 &0.84 &5.93 &0.00 \\
\cmidrule(l){2-7}
&\multirow{4}{*}{Geometric} &Scale &4.88 &0.22 &8.94 &0.00 \\
& &Rotate &27.32 &0.10 &7.30 &0.00 \\
& &Shear &24.54 &0.27 &3.31 &2.31 \\
& &Piecewse Affine &10.02 &0.25 &4.14 &4.28 \\
\end{tabular}
\end{adjustbox}
\begin{adjustbox}{width=0.60\linewidth}
\begin{tabular}{lllrrrrr}
\toprule
\textbf{Watermark} &\textbf{Attack Category} &\textbf{Perturbation} &\textbf{PSNR} &\textbf{SSIM} &\textbf{Bit Acc} &\textbf{Dect Acc} \\
\midrule
\multirow{20}{*}{RivaGAN-WM} &\multirow{4}{*}{Noise} &Gaussian &18.96 &0.40 &1.87 &12.89 \\
& &Shot &18.31 &0.41 &2.36 &12.08 \\
& &Impulse &18.18 &0.38 &4.78 &8.20 \\
& &Speckle &19.88 &0.51 &3.58 &20.71 \\
\cmidrule(l){2-7}
&\multirow{4}{*}{Blur} &Defocus &22.94 &0.63 &6.87 &32.69 \\
& &Glass &27.05 &0.83 &9.74 &84.54 \\
& &Motion &20.51 &0.61 &5.53 &32.79 \\
& &Zoom &17.32 &0.50 &3.56 &2.70 \\
\cmidrule(l){2-7}
&\multirow{4}{*}{Weather} &Snow &11.17 &0.52 &0.29 &4.38 \\
& &Frost &9.98 &0.60 &0.33 &12.97 \\
& &Fog &13.25 &0.59 &0.67 &2.37 \\
& &Brightness &14.14 &0.77 &0.85 &51.03 \\
\cmidrule(l){2-7}
&\multirow{4}{*}{Digital} &Contrast &14.74 &0.53 &0.88 &5.07 \\
& &Elastic &16.70 &0.41 &4.55 &59.62 \\
& &Pixelate &22.77 &0.65 &7.18 &42.50 \\
& &JPEG &28.07 &0.84 &5.90 &0.15 \\
\cmidrule(l){2-7}
&\multirow{4}{*}{Geometric} &Scale &15.64 &0.42 &5.82 &1.18 \\
& &Rotate &14.34 &0.23 &6.20 &72.50 \\
& &Shear &15.93 &0.68 &4.31 &5.28 \\
& &Piecewse Affine &12.21 &0.31 &2.22 &4.41 \\
\end{tabular}
\end{adjustbox}
\begin{adjustbox}{width=0.60\linewidth}
\begin{tabular}{lllrrrrr}
\toprule
\textbf{Watermark} &\textbf{Attack Category} &\textbf{Perturbation} &\textbf{PSNR} &\textbf{SSIM} &\textbf{Bit Acc} &\textbf{Dect Acc} \\
\midrule
\multirow{20}{*}{SSL-WM} &\multirow{4}{*}{Noise} &Gaussian &8.67 &0.17 &1.27 &2.50 \\
& &Shot &7.72 &0.16 &1.49 &2.43 \\
& &Impulse &7.59 &0.15 &43.87 &0.00 \\
& &Speckle &8.56 &0.22 &2.58 &13.89 \\
\cmidrule(l){2-7}
&\multirow{4}{*}{Blur} &Defocus &11.48 &0.31 &4.22 &29.46 \\
& &Glass &13.62 &0.42 &6.55 &74.71 \\
& &Motion &10.25 &0.31 &3.47 &22.36 \\
& &Zoom &8.64 &0.25 &2.08 &0.04 \\
\cmidrule(l){2-7}
&\multirow{4}{*}{Weather} &Snow &5.56 &0.26 &0.21 &3.23 \\
& &Frost &4.97 &0.30 &0.20 &29.71 \\
& &Fog &6.58 &0.30 &0.32 &59.22 \\
& &Brightness &7.04 &0.39 &1.12 &73.01 \\
\cmidrule(l){2-7}
&\multirow{4}{*}{Digital} &Contrast &7.36 &0.27 &0.49 &55.45 \\
& &Elastic &8.34 &0.21 &2.88 &46.66 \\
& &Pixelate &11.38 &0.33 &4.67 &5.60 \\
& &JPEG &14.00 &0.42 &3.19 &2.12 \\
\cmidrule(l){2-7}
&\multirow{4}{*}{Geometric} &Scale &3.77 &0.48 &26.91 &25.67 \\
& &Rotate &2.91 &0.25 &2.93 &30.82 \\
& &Shear &2.63 &0.35 &4.01 &24.19 \\
& &Piecewse Affine &3.14 &0.41 &28.53 &22.58 \\
\end{tabular}
\end{adjustbox}
\begin{adjustbox}{width=0.60\linewidth}
\begin{tabular}{lllrrrrr}
\toprule
\textbf{Watermark} &\textbf{Attack Category} &\textbf{Perturbation} &\textbf{PSNR} &\textbf{SSIM} &\textbf{Bit Acc} &\textbf{Dect Acc} \\
\midrule
\multirow{20}{*}{StegaStamp-WM} &\multirow{4}{*}{Noise} 
&Gaussian &8.41 &0.28 &2.11 &7.55 \\
& &Shot &8.33 &0.07 &2.13 &64.11 \\
& &Impulse &4.76 &0.31 &4.30 &1.80 \\
& &Speckle &8.16 &0.48 &3.25 &13.75 \\
\cmidrule(l){2-7}
&\multirow{4}{*}{Blur} &Defocus &30.26 &0.31 &3.48 &21.55 \\
& &Glass &17.98 &0.18 &3.54 &17.90 \\
& &Motion &11.48 &0.21 &3.46 &29.35 \\
& &Zoom &5.40 &0.14 &6.71 &2.70 \\
\cmidrule(l){2-7}
&\multirow{4}{*}{Weather} &Snow &6.21 &0.28 &7.38 &7.72 \\
& &Frost &4.57 &0.17 &5.73 &33.90 \\
& &Fog &4.17 &0.31 &6.07 &31.14 \\
& &Brightness &3.41 &0.28 &3.80 &1.19 \\
\cmidrule(l){2-7}
&\multirow{4}{*}{Digital} &Contrast &1.18 &0.08 &1.96 &3.88 \\
& &Elastic &3.73 &0.39 &1.87 &1.03 \\
& &Pixelate &1.95 &0.37 &2.33 &5.37 \\
& &JPEG &9.94 &0.23 &3.24 &4.80 \\
\cmidrule(l){2-7}
&\multirow{4}{*}{Geomtric} &Scale &2.33 &0.49 &3.80 &27.38 \\
& &Rotate &5.66 &0.07 &2.37 &0.27 \\
& &Shear &3.85 &0.05 &2.09 &3.41 \\
& &Piecewse Affine &4.89 &0.13 &0.10 &3.70 \\
\bottomrule
\end{tabular}
\end{adjustbox}
\end{table*}

\begin{table*}[htp]\small
\centering
\caption{Stable Diffusion image watermarks under image perturbations.}
\begin{adjustbox}{width=0.60\linewidth}
\begin{tabular}{lllrrrrr}\toprule
\textbf{Watermark} &\textbf{Attack Category} &\textbf{Perturbation} &\textbf{PSNR} &\textbf{SSIM} &\textbf{Bit Acc} &\textbf{Dect Acc} \\\midrule
\multirow{20}{*}{DctDwtSvd-WM} &\multirow{4}{*}{Noise} &Gaussian &19.12 &0.39 &1.80 &16.93 \\
& &Shot &18.62 &0.37 &2.08 &25.33 \\
& &Impulse &17.98 &0.33 &4.87 &9.81 \\
& &Speckle &20.22 &0.49 &3.73 &59.36 \\
\cmidrule(l){2-7}
&\multirow{4}{*}{Blur} &Defocus &23.93 &0.69 &11.01 &35.49 \\
& &Glass &27.61 &0.84 &15.26 &93.41 \\
& &Motion &21.20 &0.66 &7.77 &8.46 \\
& &Zoom &16.10 &0.51 &2.74 &0.00 \\
\cmidrule(l){2-7}
&\multirow{4}{*}{Weather} &Snow &11.36 &0.48 &0.37 &5.71 \\
& &Frost &16.44 &0.72 &6.68 &16.89 \\
& &Fog &12.66 &0.60 &0.55 &0.84 \\
& &Brightness &14.35 &0.76 &1.00 &19.27 \\
\cmidrule(l){2-7}
&\multirow{4}{*}{Digital} &Contrast &17.50 &0.55 &0.58 &1.70 \\
& &Elastic &21.05 &0.49 &6.94 &14.85 \\
& &Pixelate &22.03 &0.55 &8.06 &0.56 \\
& &JPEG &27.83 &0.82 &6.29 &0.00 \\
\cmidrule(l){2-7}
&\multirow{4}{*}{Geometric} &Scale &11.04 &0.23 &1.72 &0.00 \\
& &Rotate &9.23 &0.17 &1.27 &0.00 \\
& &Shear &10.96 &0.24 &2.03 &0.00 \\
& &Piecewse Affine &14.48 &0.36 &3.78 &7.23 \\
\end{tabular}
\end{adjustbox}
\begin{adjustbox}{width=0.60\linewidth}
\begin{tabular}{lllrrrrr}
\toprule
\textbf{Watermark} &\textbf{Attack Category} &\textbf{Perturbation} &\textbf{PSNR} &\textbf{SSIM} &\textbf{Bit Acc} &\textbf{Dect Acc} \\
\midrule
\multirow{20}{*}{RivaGAN-WM} &\multirow{4}{*}{Noise} &Gaussian &19.03 &0.35 &1.78 &12.69 \\
& &Shot &18.56 &0.37 &2.00 &15.65 \\
& &Impulse &17.93 &0.33 &4.10 &6.36 \\
& &Speckle &20.14 &0.49 &3.07 &29.42 \\
\cmidrule(l){2-7}
&\multirow{4}{*}{Blur} &Defocus &23.81 &0.69 &7.00 &46.98 \\
& &Glass &27.23 &0.82 &7.91 &95.22 \\
& &Motion &21.14 &0.65 &5.34 &47.78 \\
& &Zoom &16.09 &0.51 &2.36 &0.60 \\
\cmidrule(l){2-7}
&\multirow{4}{*}{Weather} &Snow &11.38 &0.48 &0.37 &6.85 \\
& &Frost &15.47 &0.71 &1.53 &46.23 \\
& &Fog &12.58 &0.60 &0.56 &8.27 \\
& &Brightness &14.39 &0.73 &0.76 &48.22 \\
\cmidrule(l){2-7}
&\multirow{4}{*}{Digital} &Contrast &14.06 &0.54 &0.58 &8.23 \\
& &Elastic &17.46 &0.49 &3.62 &84.18 \\
& &Pixelate &21.01 &0.55 &5.51 &4.17 \\
& &JPEG &27.71 &0.81 &5.78 &2.26 \\
\cmidrule(l){2-7}
&\multirow{4}{*}{Geometric} &Scale &11.04 &0.23 &1.79 &33.00 \\
& &Rotate &9.23 &0.17 &1.30 &13.00 \\
& &Shear &10.96 &0.24 &2.31 &60.28 \\
& &Piecewse Affine &14.48 &0.35 &4.24 &58.45 \\
\end{tabular}
\end{adjustbox}
\begin{adjustbox}{width=0.60\linewidth}
\begin{tabular}{lllrrrrr}
\toprule
\textbf{Watermark} &\textbf{Attack Category} &\textbf{Perturbation} &\textbf{PSNR} &\textbf{SSIM} &\textbf{Bit Acc} &\textbf{Dect Acc} \\
\midrule
\multirow{20}{*}{SSL-WM} &\multirow{4}{*}{Noise} &Gaussian &9.12 &0.33 &1.31 &0.00 \\
& &Shot &6.61 &0.21 &2.26 &0.00 \\
& &Impulse &7.03 &0.51 &4.83 &0.00 \\
& &Speckle &4.15 &0.43 &3.33 &3.41 \\
\cmidrule(l){2-7}
&\multirow{4}{*}{Blur} &Defocus &22.78 &0.33 &3.98 &28.31 \\
& &Glass &18.67 &0.28 &2.19 &88.36 \\
& &Motion &10.00 &0.18 &3.21 &45.87 \\
& &Zoom &9.53 &0.27 &1.19 &4.86 \\
\cmidrule(l){2-7}
&\multirow{4}{*}{Weather} &Snow &3.64 &0.15 &0.80 &5.70 \\
& &Frost &4.86 &0.22 &0.78 &18.43 \\
& &Fog &5.94 &0.32 &1.11 &39.85 \\
& &Brightness &6.93 &0.33 &1.41 &41.83 \\
\cmidrule(l){2-7}
&\multirow{4}{*}{Digital} &Contrast &6.92 &0.38 &2.85 &65.78 \\
& &Elastic &5.23 &0.40 &3.95 &33.84 \\
& &Pixelate &8.82 &0.21 &1.31 &21.73 \\
& &JPEG &7.98 &0.38 &4.43 &22.38 \\
\cmidrule(l){2-7}
&\multirow{4}{*}{Geometric} &Scale &5.79 &0.38 &2.36 &24.46 \\
& &Rotate &6.04 &0.25 &3.17 &25.90 \\
& &Shear &5.35 &0.38 &2.80 &23.78 \\
& &Piecewse Affine &6.28 &0.18 &2.88 &0.00 \\
\end{tabular}
\end{adjustbox}
\begin{adjustbox}{width=0.60\linewidth}
\begin{tabular}{lllrrrrr}
\toprule
\textbf{Watermark} &\textbf{Attack Category} &\textbf{Perturbation} &\textbf{PSNR} &\textbf{SSIM} &\textbf{Bit Acc} &\textbf{Dect Acc} \\
\midrule
\multirow{20}{*}{StegaStamp-WM} &\multirow{4}{*}{Noise} &Gaussian &8.75 &0.18 &2.84 &0.00 \\
& &Shot &7.03 &0.40 &3.90 &0.46 \\
& &Impulse &6.31 &0.38 &3.34 &0.37 \\
& &Speckle &7.73 &0.34 &2.45 &22.85 \\
\cmidrule(l){2-7}
&\multirow{4}{*}{Blur} &Defocus &32.87 &0.28 &2.67 &24.96 \\
& &Glass &19.54 &0.26 &3.86 &21.14 \\
& &Motion &11.48 &0.18 &4.83 &23.84 \\
& &Zoom &4.95 &0.00 &5.28 &14.80 \\
\cmidrule(l){2-7}
&\multirow{4}{*}{Weather} &Snow &6.76 &0.43 &6.99 &33.57 \\
& &Frost &4.05 &0.36 &6.78 &35.76 \\
& &Fog &6.13 &0.29 &6.57 &24.05 \\
& &Brightness &1.23 &0.35 &6.47 &0.00 \\
\cmidrule(l){2-7}
&\multirow{4}{*}{Digital} &Contrast &1.13 &0.22 &6.78 &0.00 \\
& &Elastic &2.08 &0.42 &5.72 &3.72 \\
& &Pixelate &1.93 &0.37 &1.74 &25.97 \\
& &JPEG &9.28 &0.25 &1.85 &33.78 \\
\cmidrule(l){2-7}
&\multirow{4}{*}{Geomtric} &Scale &3.50 &0.33 &2.04 &26.58 \\
& &Rotate &4.82 &0.37 &2.38 &38.56 \\
& &Shear &3.81 &0.22 &2.17 &27.31 \\
& &Piecewse Affine &4.22 &0.21 &2.10 &27.44 \\
\bottomrule
\end{tabular}
\end{adjustbox}
\end{table*}

\begin{table*}[htp]\small
\centering
\caption{DALLE3 image watermarks under image perturbations.}
\begin{adjustbox}{width=0.60\linewidth}
\begin{tabular}{lllrrrrr}\toprule
\textbf{Watermark} &\textbf{Attack Category} &\textbf{Perturbation} &\textbf{PSNR} &\textbf{SSIM} &\textbf{Bit Acc} &\textbf{Dect Acc} \\\midrule
\multirow{20}{*}{DctDwtSvd-WM} &\multirow{4}{*}{Noise} &Gaussian &18.91 &0.39 &1.83 &12.61 \\
& &Shot &18.15 &0.39 &2.22 &16.81 \\
& &Impulse &18.18 &0.37 &4.42 &7.38 \\
& &Speckle &19.62 &0.48 &3.05 &39.09 \\
\cmidrule(l){2-7}
&\multirow{4}{*}{Blur} &Defocus &22.90 &0.64 &5.64 &33.88 \\
& &Glass &27.26 &0.84 &7.83 &88.00 \\
& &Motion &20.30 &0.62 &5.10 &9.63 \\
& &Zoom &16.18 &0.49 &2.83 &0.00 \\
\cmidrule(l){2-7}
&\multirow{4}{*}{Weather} &Snow &11.13 &0.51 &0.31 &10.75 \\
& &Frost &10.36 &0.60 &0.40 &5.13 \\
& &Fog &13.21 &0.61 &0.65 &1.08 \\
& &Brightness &13.84 &0.77 &0.78 &28.58 \\
\cmidrule(l){2-7}
&\multirow{4}{*}{Digital} &Contrast &14.48 &0.54 &0.83 &2.40 \\
& &Elastic &16.50 &0.43 &4.20 &17.17 \\
& &Pixelate &21.04 &0.56 &5.23 &17.97 \\
& &JPEG &28.63 &0.85 &5.89 &0.00 \\
\cmidrule(l){2-7}
&\multirow{4}{*}{Geomtric} &Scale &10.88 &0.25 &1.50 &9.98 \\
& &Rotate &9.31 &0.19 &1.09 &8.51 \\
& &Shear &10.72 &0.26 &2.40 &4.15 \\
& &Piecewse Affine &15.50 &0.43 &5.58 &3.09 \\
\end{tabular}
\end{adjustbox}
\begin{adjustbox}{width=0.60\linewidth}
\begin{tabular}{lllrrrrr}
\toprule
\textbf{Watermark} &\textbf{Attack Category} &\textbf{Perturbation} &\textbf{PSNR} &\textbf{SSIM} &\textbf{Bit Acc} &\textbf{Dect Acc} \\
\midrule
\multirow{20}{*}{RivaGAN-WM} &\multirow{4}{*}{Noise} &Gaussian &18.97 &0.39 &1.86 &15.96 \\
& &Shot &18.21 &0.40 &2.33 &14.4 \\
& &Impulse &18.22 &0.37 &5.06 &11.07 \\
& &Speckle &19.71 &0.49 &3.71 &23.89 \\
\cmidrule(l){2-7}
&\multirow{4}{*}{Blur} &Defocus &23.02 &0.65 &7.86 &37.77 \\
& &Glass &27.64 &0.86 &12.50 &91.53 \\
& &Motion &20.34 &0.62 &6.30 &38.05 \\
& &Zoom &16.19 &0.49 &3.19 &0.6 \\
\cmidrule(l){2-7}
&\multirow{4}{*}{Weather} &Snow &11.17 &0.52 &0.32 &5.9 \\
& &Frost &10.40 &0.61 &0.31 &14.36 \\
& &Fog &13.22 &0.62 &0.40 &4.22 \\
& &Brightness &13.98 &0.79 &0.65 &56.86 \\
\cmidrule(l){2-7}
&\multirow{4}{*}{Digital} &Contrast &14.49 &0.55 &0.99 &6.47 \\
& &Elastic &16.54 &0.43 &5.17 &74.38 \\
& &Pixelate &21.10 &0.57 &6.72 &20.17 \\
& &JPEG &28.82 &0.85 &6.20 &0.63 \\
\cmidrule(l){2-7}
&\multirow{4}{*}{Geomtric} &Scale &10.87 &0.25 &1.36 &75.57 \\
& &Rotate &9.31 &0.18 &1.00 &16.16 \\
& &Shear &10.81 &0.27 &1.81 &0.00 \\
& &Piecewse Affine &15.57 &0.42 &4.02 &0.00 \\
\end{tabular}
\end{adjustbox}
\begin{adjustbox}{width=0.60\linewidth}
\begin{tabular}{lllrrrrr}
\toprule
\textbf{Watermark} &\textbf{Attack Category} &\textbf{Perturbation} &\textbf{PSNR} &\textbf{SSIM} &\textbf{Bit Acc} &\textbf{Dect Acc} \\
\midrule
\multirow{20}{*}{SSL-WM} &\multirow{4}{*}{Noise} &Gaussian &10.77 &0.02 &0.08 &3.41 \\
& &Shot &11.83 &0.08 &3.31 &2.33 \\
& &Impulse &5.58 &0.19 &3.76 &4.82 \\
& &Speckle &6.93 &0.35 &3.58 &1.83 \\
\cmidrule(l){2-7}
&\multirow{4}{*}{Blur} &Defocus &22.73 &0.28 &3.17 &18.88 \\
& &Glass &25.48 &0.58 &3.21 &66.63 \\
& &Motion &21.85 &0.46 &4.75 &25.34 \\
& &Zoom &18.42 &0.31 &4.24 &28.64 \\
\cmidrule(l){2-7}
&\multirow{4}{*}{Weather} &Snow &4.31 &0.27 &3.82 &27.80 \\
& &Frost &5.68 &0.28 &1.41 &47.85 \\
& &Fog &8.80 &0.35 &3.89 &44.42 \\
& &Brightness &9.41 &0.34 &5.91 &41.23 \\
\cmidrule(l){2-7}
&\multirow{4}{*}{Digital} &Contrast &10.53 &0.30 &1.23 &37.64 \\
& &Elastic &11.32 &0.38 &1.35 &36.94 \\
& &Pixelate &6.42 &0.32 &2.94 &21.44 \\
& &JPEG &3.31 &0.35 &3.74 &22.85 \\
\cmidrule(l){2-7}
&\multirow{4}{*}{Geomtric} &Scale &5.83 &0.14 &9.44 &38.74 \\
& &Rotate &2.78 &0.32 &1.42 &14.84 \\
& &Shear &3.33 &0.02 &2.83 &24.00 \\
& &Piecewse Affine &2.82 &0.01 &8.27 &3.78 \\
\end{tabular}
\end{adjustbox}
\begin{adjustbox}{width=0.60\linewidth}
\begin{tabular}{lllrrrrr}
\toprule
\textbf{Watermark} &\textbf{Attack Category} &\textbf{Perturbation} &\textbf{PSNR} &\textbf{SSIM} &\textbf{Bit Acc} &\textbf{Dect Acc} \\
\midrule
\multirow{20}{*}{StegaStamp-WM} &\multirow{4}{*}{Noise} &Gaussian &5.02 &0.24 &1.49 &54.41 \\
& &Shot &3.95 &0.41 &2.41 &55.31 \\
& &Impulse &7.33 &0.11 &2.01 &2.25 \\
& &Speckle &8.04 &0.29 &3.82 &9.62 \\
\cmidrule(l){2-7}
&\multirow{4}{*}{Blur} &Defocus &8.75 &0.20 &6.93 &17.83 \\
& &Glass &12.88 &0.38 &4.92 &14.89 \\
& &Motion &13.85 &0.10 &6.21 &5.04 \\
& &Zoom &13.00 &0.17 &4.82 &1.73 \\
\cmidrule(l){2-7}
&\multirow{4}{*}{Weather} &Snow &1.65 &0.39 &4.88 &10.85 \\
& &Frost &3.28 &0.28 &6.28 &29.59 \\
& &Fog &3.90 &0.12 &2.08 &18.13 \\
& &Brightness &4.00 &0.18 &5.32 &19.41 \\
\cmidrule(l){2-7}
&\multirow{4}{*}{Digital} &Contrast &8.53 &0.42 &0.53 &27.74 \\
& &Elastic &20.41 &0.40 &3.88 &24.93 \\
& &Pixelate &3.45 &0.43 &5.83 &1.07 \\
& &JPEG &6.80 &0.54 &2.97 &8.10 \\
\cmidrule(l){2-7}
&\multirow{4}{*}{Geomtric} &Scale &2.58 &0.18 &1.25 &3.31 \\
& &Rotate &2.87 &0.28 &2.28 &3.91 \\
& &Shear &3.95 &0.27 &5.12 &8.00 \\
& &Piecewse Affine &2.37 &0.39 &3.76 &3.41 \\
\bottomrule
\end{tabular}
\end{adjustbox}
\end{table*}

\begin{table*}[htp]\small
\centering
\caption{SDXL-Lightning image watermarks under image perturbations.}
\begin{adjustbox}{width=0.60\linewidth}
\begin{tabular}{lllrrrrr}\toprule
\textbf{Watermark} &\textbf{Attack Category} &\textbf{Perturbation} &\textbf{PSNR} &\textbf{SSIM} &\textbf{Bit Acc} &\textbf{Dect Acc} \\\midrule
\multirow{20}{*}{DctDwtSvd-WM} &\multirow{4}{*}{Noise} &Gaussian &18.90 &0.29 &1.55 &19.75 \\
& &Shot &18.08 &0.28 &1.52 &22.86 \\
& &Impulse &18.36 &0.27 &4.53 &17.43 \\
& &Speckle &19.73 &0.38 &2.50 &51.08 \\
\cmidrule(l){2-7}
&\multirow{4}{*}{Blur} &Defocus &26.79 &0.79 &17.35 &38.88 \\
& &Glass &31.14 &0.91 &23.44 &96.75 \\
& &Motion &23.46 &0.74 &11.21 &14.25 \\
& &Zoom &17.83 &0.61 &3.65 &0.00 \\
\cmidrule(l){2-7}
&\multirow{4}{*}{Weather} &Snow &11.14 &0.49 &0.03 &16.47 \\
& &Frost &16.97 &0.72 &7.08 &25.29 \\
& &Fog &13.71 &0.69 &0.69 &0.80 \\
& &Brightness &13.52 &0.81 &0.07 &29.25 \\
\cmidrule(l){2-7}
&\multirow{4}{*}{Digital} &Contrast &15.68 &0.65 &0.85 &3.10 \\
& &Elastic &19.63 &0.58 &10.63 &26.39 \\
& &Pixelate &23.30 &0.65 &12.28 &7.29 \\
& &JPEG &30.64 &0.87 &7.62 &0.00 \\
\cmidrule(l){2-7}
&\multirow{4}{*}{Geomtric} &Scale &12.46 &0.35 &1.81 &22.00 \\
& &Rotate &10.22 &0.27 &1.27 &23.56 \\
& &Shear &12.37 &0.36 &3.17 &0.00 \\
& &Piecewse Affine &17.06 &0.52 &8.02 &10.17 \\
\end{tabular}
\end{adjustbox}
\begin{adjustbox}{width=0.60\linewidth}
\begin{tabular}{lllrrrrr}
\toprule
\textbf{Watermark} &\textbf{Attack Category} &\textbf{Perturbation} &\textbf{PSNR} &\textbf{SSIM} &\textbf{Bit Acc} &\textbf{Dect Acc} \\
\midrule
\multirow{20}{*}{RivaGAN-WM} &\multirow{4}{*}{Noise} &Gaussian &18.84 &0.28 &1.53 &9.11 \\
& &Shot &18.04 &0.28 &1.50 &9.60 \\
& &Impulse &18.31 &0.27 &3.92 &5.23 \\
& &Speckle &19.68 &0.37 &2.32 &19.15 \\
\cmidrule(l){2-7}
&\multirow{4}{*}{Blur} &Defocus &26.55 &0.77 &9.33 &51.98 \\
& &Glass &30.33 &0.89 &9.78 &98.13 \\
& &Motion &23.35 &0.74 &6.69 &52.41 \\
& &Zoom &17.82 &0.60 &3.18 &0.98 \\
\cmidrule(l){2-7}
&\multirow{4}{*}{Weather} &Snow &11.18 &0.48 &0.03 &9.41 \\
& &Frost &15.82 &0.71 &1.35 &46.92 \\
& &Fog &13.68 &0.69 &0.69 &6.58 \\
& &Brightness &13.62 &0.77 &0.08 &54.60 \\
\cmidrule(l){2-7}
&\multirow{4}{*}{Digital} &Contrast &15.67 &0.65 &0.85 &6.15 \\
& &Elastic &19.54 &0.57 &5.72 &90.94 \\
& &Pixelate &23.22 &0.64 &7.49 &8.86 \\
& &JPEG &30.28 &0.86 &6.95 &1.902 \\
\cmidrule(l){2-7}
&\multirow{4}{*}{Geomtric} &Scale &12.47 &0.34 &1.64 &82.98 \\
& &Rotate &10.22 &0.26 &1.15 &16.29 \\
& &Shear &12.37 &0.35 &2.13 &76.05 \\
& &Piecewse Affine &17.03 &0.51 &4.86 &89.35 \\
\end{tabular}
\end{adjustbox}
\begin{adjustbox}{width=0.60\linewidth}
\begin{tabular}{lllrrrrr}
\toprule
\textbf{Watermark} &\textbf{Attack Category} &\textbf{Perturbation} &\textbf{PSNR} &\textbf{SSIM} &\textbf{Bit Acc} &\textbf{Dect Acc} \\
\midrule
\multirow{20}{*}{SSL-WM} &\multirow{4}{*}{Noise} &Gaussian &11.24 &0.05 &0.10 &3.75 \\
& &Shot &12.45 &0.12 &3.62 &2.67 \\
& &Impulse &6.12 &0.24 &3.87 &5.16 \\
& &Speckle &7.56 &0.39 &3.69 &2.18 \\
\cmidrule(l){2-7}
&\multirow{4}{*}{Blur} &Defocus &23.11 &0.31 &3.28 &30.33 \\
& &Glass &26.02 &0.55 &3.32 &69.81 \\
& &Motion &22.28 &0.48 &4.86 &27.75 \\
& &Zoom &19.05 &0.33 &4.35 &31.06 \\
\cmidrule(l){2-7}
&\multirow{4}{*}{Weather} &Snow &4.89 &0.29 &3.93 &29.22 \\
& &Frost &6.17 &0.30 &1.61 &49.27 \\
& &Fog &9.34 &0.37 &3.90 &45.84 \\
& &Brightness &10.03 &0.36 &6.02 &42.65 \\
\cmidrule(l){2-7}
&\multirow{4}{*}{Digital} &Contrast &11.15 &0.32 &1.43 &38.06 \\
& &Elastic &12.00 &0.41 &1.55 &37.36 \\
& &Pixelate &7.01 &0.34 &3.04 &23.87 \\
& &JPEG &3.87 &0.37 &3.84 &25.28 \\
\cmidrule(l){2-7}
&\multirow{4}{*}{Geomtric} &Scale &4.75 &0.31 &5.88 &24.95 \\
& &Rotate &6.27 &0.28 &4.72 &24.24 \\
& &Shear &5.43 &0.30 &4.70 &28.57 \\
& &Piecewse Affine &4.22 &0.28 &4.71 &20.87 \\
\end{tabular}
\end{adjustbox}
\begin{adjustbox}{width=0.60\linewidth}
\begin{tabular}{lllrrrrr}
\toprule
\textbf{Watermark} &\textbf{Attack Category} &\textbf{Perturbation} &\textbf{PSNR} &\textbf{SSIM} &\textbf{Bit Acc} &\textbf{Dect Acc} \\
\midrule
\multirow{20}{*}{StegaStamp-WM} &\multirow{4}{*}{Noise} &Gaussian &5.19 &0.53 &9.75 &9.11 \\
& &Shot &3.85 &0.05 &2.57 &2.84 \\
& &Impulse &7.25 &0.17 &1.87 &9.92 \\
& &Speckle &8.61 &0.24 &3.95 &14.80 \\
\cmidrule(l){2-7}
&\multirow{4}{*}{Blur} &Defocus &6.48 &0.28 &7.01 &1.20 \\
& &Glass &12.11 &0.38 &5.02 &1.17 \\
& &Motion &9.03 &0.31 &4.18 &3.10 \\
& &Zoom &14.01 &0.17 &4.38 &0.27 \\
\cmidrule(l){2-7}
&\multirow{4}{*}{Weather} &Snow &2.78 &0.19 &6.28 &1.50 \\
& &Frost &3.48 &0.02 &6.82 &14.04 \\
& &Fog &4.13 &0.03 &2.47 &29.84 \\
& &Brightness &3.02 &0.20 &5.73 &24.48 \\
\cmidrule(l){2-7}
&\multirow{4}{*}{Digital} &Contrast &12.41 &0.21 &7.47 &24.79 \\
& &Elastic &16.82 &0.15 &2.75 &21.28 \\
& &Pixelate &3.57 &0.24 &4.72 &5.34 \\
& &JPEG &6.63 &0.22 &5.37 &4.32 \\
\cmidrule(l){2-7}
&\multirow{4}{*}{Geomtric} &Scale &2.73 &0.38 &3.88 &3.57 \\
& &Rotate &4.93 &0.42 &1.57 &5.16 \\
& &Shear &2.54 &0.21 &4.38 &6.28 \\
& &Piecewse Affine &2.44 &0.10 &5.07 &2.04 \\
\bottomrule
\end{tabular}
\end{adjustbox}
\end{table*}

\begin{table*}[htp]\small
\centering
\caption{PIXART image watermarks under image perturbations.}
\begin{adjustbox}{width=0.60\linewidth}
\begin{tabular}{lllrrrrr}\toprule
\textbf{Watermark} &\textbf{Attack Category} &\textbf{Perturbation} &\textbf{PSNR} &\textbf{SSIM} &\textbf{Bit Acc} &\textbf{Dect Acc} \\\midrule
\multirow{20}{*}{DctDwtSvd-WM} &\multirow{4}{*}{Noise} &Gaussian &19.19 &0.30 &2.23 &14.15 \\
& &Shot &19.23 &0.36 &3.40 &35.73 \\
& &Impulse &17.70 &0.27 &5.01 &8.26 \\
& &Speckle &20.87 &0.49 &4.75 &67.80 \\
\cmidrule(l){2-7}
&\multirow{4}{*}{Blur} &Defocus &24.43 &0.75 &8.24 &38.18 \\
& &Glass &27.98 &0.86 &10.37 &92.87 \\
& &Motion &21.89 &0.71 &7.40 &9.56 \\
& &Zoom &16.86 &0.57 &3.47 &0.00 \\
\cmidrule(l){2-7}
&\multirow{4}{*}{Weather} &Snow &11.45 &0.43 &0.49 &10.60 \\
& &Frost &15.81 &0.66 &2.68 &19.44 \\
& &Fog &11.88 &0.57 &0.47 &0.72 \\
& &Brightness &14.77 &0.70 &1.69 &18.32 \\
\cmidrule(l){2-7}
&\multirow{4}{*}{Digital} &Contrast &13.94 &0.56 &0.61 &2.10 \\
& &Elastic &18.28 &0.57 &6.32 &21.72 \\
& &Pixelate &21.65 &0.63 &7.07 &5.11 \\
& &JPEG &28.70 &0.83 &7.23 &0.00 \\
\cmidrule(l){2-7}
&\multirow{4}{*}{Geomtric} &Scale &11.36 &0.33 &2.26 &1.43 \\
& &Rotate &9.83 &0.26 &2.06 &16.60 \\
& &Shear &11.18 &0.35 &3.05 &6.49 \\
& &Piecewse Affine &16.28 &0.51 &6.80 &7.56 \\
\end{tabular}
\end{adjustbox}
\begin{adjustbox}{width=0.60\linewidth}
\begin{tabular}{lllrrrrr}
\toprule
\textbf{Watermark} &\textbf{Attack Category} &\textbf{Perturbation} &\textbf{PSNR} &\textbf{SSIM} &\textbf{Bit Acc} &\textbf{Dect Acc} \\
\midrule
\multirow{20}{*}{RivaGAN-WM} &\multirow{4}{*}{Noise} &Gaussian &19.25 &0.30 &2.27 &5.58 \\
& &Shot &19.31 &0.37 &3.79 &13.11 \\
& &Impulse &17.73 &0.27 &5.76 &1.81 \\
& &Speckle &20.98 &0.51 &7.90 &28.58 \\
\cmidrule(l){2-7}
&\multirow{4}{*}{Blur} &Defocus &24.58 &0.76 &16.53 &47.98 \\
& &Glass &28.39 &0.89 &22.61 &94.15 \\
& &Motion &21.96 &0.73 &12.57 &48.42 \\
& &Zoom &16.87 &0.58 &5.23 &0.47 \\
\cmidrule(l){2-7}
&\multirow{4}{*}{Weather} &Snow &11.50 &0.44 &0.50 &5.56 \\
& &Frost &16.50 &0.67 &7.89 &42.01 \\
& &Fog &11.88 &0.57 &0.46 &4.85 \\
& &Brightness &14.97 &0.73 &2.27 &31.75 \\
\cmidrule(l){2-7}
&\multirow{4}{*}{Digital} &Contrast &13.94 &0.56 &0.61 &5.60 \\
& &Elastic &18.30 &0.58 &11.13 &82.63 \\
& &Pixelate &21.70 &0.64 &12.52 &5.54 \\
& &JPEG &28.93 &0.84 &7.92 &2.21 \\
\cmidrule(l){2-7}
&\multirow{4}{*}{Geomtric} &Scale &11.36 &0.33 &3.15 &74.49 \\
& &Rotate &9.83 &0.26 &2.73 &15.30 \\
& &Shear &11.18 &0.35 &4.96 &69.68 \\
& &Piecewse Affine &17.27 &0.52 &9.30 &82.23 \\
\end{tabular}
\end{adjustbox}
\begin{adjustbox}{width=0.60\linewidth}
\begin{tabular}{lllrrrrr}
\toprule
\textbf{Watermark} &\textbf{Attack Category} &\textbf{Perturbation} &\textbf{PSNR} &\textbf{SSIM} &\textbf{Bit Acc} &\textbf{Dect Acc} \\
\midrule
\multirow{20}{*}{SSL-WM} &\multirow{4}{*}{Noise} &Gaussian &9.56 &0.35 &1.43 &0.31 \\
& &Shot &7.07 &0.23 &2.38 &0.99 \\
& &Impulse &7.49 &0.52 &4.95 &0.45 \\
& &Speckle &4.31 &0.44 &3.45 &3.61 \\
\cmidrule(l){2-7}
&\multirow{4}{*}{Blur} &Defocus &23.04 &0.34 &4.10 &39.60 \\
& &Glass &19.93 &0.29 &2.21 &89.56 \\
& &Motion &11.30 &0.19 &3.23 &46.95 \\
& &Zoom &10.83 &0.27 &1.29 &5.16 \\
\cmidrule(l){2-7}
&\multirow{4}{*}{Weather} &Snow &4.14 &0.16 &0.82 &6.00 \\
& &Frost &5.36 &0.23 &0.80 &19.53 \\
& &Fog &6.44 &0.33 &1.13 &39.45 \\
& &Brightness &7.43 &0.34 &1.43 &42.13 \\
\cmidrule(l){2-7}
&\multirow{4}{*}{Digital} &Contrast &7.42 &0.49 &2.87 &55.06 \\
& &Elastic &5.73 &0.41 &3.98 &34.14 \\
& &Pixelate &9.02 &0.22 &1.33 &22.93 \\
& &JPEG &8.18 &0.39 &4.45 &23.56 \\
\cmidrule(l){2-7}
&\multirow{4}{*}{Geomtric} &Scale &9.90 &0.02 &4.04 &23.46 \\
& &Rotate &7.65 &0.37 &5.73 &22.74 \\
& &Shear &2.64 &0.28 &6.79 &21.57 \\
& &Piecewse Affine &3.35 &0.21 &5.46 &0.00 \\
\end{tabular}
\end{adjustbox}
\begin{adjustbox}{width=0.60\linewidth}
\begin{tabular}{lllrrrrr}
\toprule
\textbf{Watermark} &\textbf{Attack Category} &\textbf{Perturbation} &\textbf{PSNR} &\textbf{SSIM} &\textbf{Bit Acc} &\textbf{Dect Acc} \\
\midrule
\multirow{20}{*}{StegaStamp-WM} &\multirow{4}{*}{Noise} &Gaussian &4.07 &0.24 &2.11 &21.84 \\
& &Shot &4.47 &0.53 &2.49 &6.72 \\
& &Impulse &7.58 &0.24 &1.92 &14.95 \\
& &Speckle &8.58 &0.16 &3.00 &18.75 \\
\cmidrule(l){2-7}
&\multirow{4}{*}{Blur} &Defocus &10.33 &0.39 &4.82 &21.75 \\
& &Glass &11.74 &0.05 &5.08 &22.92 \\
& &Motion &13.75 &0.42 &5.71 &10.32 \\
& &Zoom &10.99 &0.00 &4.32 &20.30 \\
\cmidrule(l){2-7}
&\multirow{4}{*}{Weather} &Snow &10.41 &0.45 &5.55 &44.78 \\
& &Frost &17.04 &0.37 &4.18 &17.81 \\
& &Fog &3.56 &0.03 &2.25 &14.75 \\
& &Brightness &4.08 &0.10 &5.14 &21.80 \\
\cmidrule(l){2-7}
&\multirow{4}{*}{Digital} &Contrast &13.77 &0.09 &0.05 &23.72 \\
& &Elastic &18.80 &0.22 &2.10 &3.11 \\
& &Pixelate &3.53 &0.21 &3.04 &1.82 \\
& &JPEG &6.10 &0.30 &5.28 &9.72 \\
\cmidrule(l){2-7}
&\multirow{4}{*}{Geomtric} &Scale &7.75 &0.04 &3.48 &1.85 \\
& &Rotate &1.78 &0.09 &0.24 &5.60 \\
& &Shear &4.37 &0.05 &2.93 &4.42 \\
& &Piecewse Affine &0.28 &0.13 &2.47 &3.72 \\
\bottomrule
\end{tabular}
\end{adjustbox}
\end{table*}

\begin{table*}[htp]\small
\centering
\caption{Kandinsky 2.2 image watermarks under image perturbations.}
\begin{adjustbox}{width=0.60\linewidth}
\begin{tabular}{lllrrrrr}\toprule
\textbf{Watermark} &\textbf{Attack Category} &\textbf{Perturbation} &\textbf{PSNR} &\textbf{SSIM} &\textbf{Bit Acc} &\textbf{Dect Acc} \\\midrule
\multirow{20}{*}{DctDwtSvd-WM} &\multirow{4}{*}{Noise} &Gaussian &19.25 &0.46 &2.90 &1.20 \\
& &Shot &18.71 &0.49 &4.41 &7.82 \\
& &Impulse &17.55 &0.42 &6.01 &0.12 \\
& &Speckle &19.64 &0.58 &5.44 &22.25 \\
\cmidrule(l){2-7}
&\multirow{4}{*}{Blur} &Defocus &21.15 &0.64 &5.96 &19.12 \\
& &Glass &25.33 &0.83 &8.45 &58.70 \\
& &Motion &18.63 &0.60 &5.44 &2.37 \\
& &Zoom &15.09 &0.45 &3.27 &0.00 \\
\cmidrule(l){2-7}
&\multirow{4}{*}{Weather} &Snow &11.65 &0.53 &1.46 &0.66 \\
& &Frost &10.25 &0.54 &1.47 &0.40 \\
& &Fog &11.27 &0.50 &0.39 &0.08 \\
& &Brightness &15.85 &0.72 &3.88 &10.37 \\
\cmidrule(l){2-7}
&\multirow{4}{*}{Digital} &Contrast &12.48 &0.44 &0.41 &0.72 \\
& &Elastic &14.65 &0.40 &4.78 &4.81 \\
& &Pixelate &20.95 &0.65 &6.32 &20.45 \\
& &JPEG &26.83 &0.84 &6.89 &0.00 \\
\cmidrule(l){2-7}
&\multirow{4}{*}{Geomtric} &Scale &9.85 &0.19 &2.57 &0.00 \\
& &Rotate &8.15 &0.14 &2.50 &0.00 \\
& &Shear &9.91 &0.23 &3.20 &0.00 \\
& &Piecewse Affine &12.70 &0.32 &4.43 &1.13 \\
\end{tabular}
\end{adjustbox}
\begin{adjustbox}{width=0.60\linewidth}
\begin{tabular}{lllrrrrr}
\toprule
\textbf{Watermark} &\textbf{Attack Category} &\textbf{Perturbation} &\textbf{PSNR} &\textbf{SSIM} &\textbf{Bit Acc} &\textbf{Dect Acc} \\
\midrule
\multirow{20}{*}{RivaGAN-WM} &\multirow{4}{*}{Noise} &Gaussian &19.31 &0.46 &3.00 &7.05 \\
& &Shot &18.77 &0.50 &4.78 &9.32 \\
& &Impulse &17.59 &0.42 &6.82 &3.17 \\
& &Speckle &19.72 &0.59 &7.89 &16.15 \\
\cmidrule(l){2-7}
&\multirow{4}{*}{Blur} &Defocus &21.23 &0.65 &9.64 &25.81 \\
& &Glass &25.57 &0.85 &14.67 &68.94 \\
& &Motion &18.65 &0.61 &7.86 &24.75 \\
& &Zoom &15.10 &0.46 &4.76 &1.90 \\
\cmidrule(l){2-7}
&\multirow{4}{*}{Weather} &Snow &11.69 &0.53 &1.48 &2.33 \\
& &Frost &10.29 &0.55 &1.47 &5.24 \\
& &Fog &11.29 &0.50 &0.40 &1.13 \\
& &Brightness &16.05 &0.74 &5.24 &32.11 \\
\cmidrule(l){2-7}
&\multirow{4}{*}{Digital} &Contrast &12.49 &0.44 &0.41 &3.43 \\
& &Elastic &14.65 &0.40 &7.17 &40.7 \\
& &Pixelate &21.02 &0.67 &10.42 &32.35 \\
& &JPEG &27.00 &0.85 &7.29 &0.4 \\
\cmidrule(l){2-7}
&\multirow{4}{*}{Geomtric} &Scale &9.86 &0.19 &3.36 &3.86 \\
& &Rotate &8.15 &0.14 &3.03 &4.85 \\
& &Shear &9.91 &0.23 &4.68 &40.43 \\
& &Piecewse Affine &12.69 &0.33 &6.13 &19.56 \\
\end{tabular}
\end{adjustbox}
\begin{adjustbox}{width=0.60\linewidth}
\begin{tabular}{lllrrrrr}
\toprule
\textbf{Watermark} &\textbf{Attack Category} &\textbf{Perturbation} &\textbf{PSNR} &\textbf{SSIM} &\textbf{Bit Acc} &\textbf{Dect Acc} \\
\midrule
\multirow{20}{*}{SSL-WM} &\multirow{4}{*}{Noise} &Gaussian &5.82 &0.15 &1.19 &1.11 \\
& &Shot &6.88 &0.22 &2.83 &1.41 \\
& &Impulse &7.31 &0.31 &3.33 &21.34 \\
& &Speckle &7.82 &0.10 &6.21 &26.59 \\
\cmidrule(l){2-7}
&\multirow{4}{*}{Blur} &Defocus &7.95 &0.11 &4.31 &27.46 \\
& &Glass &10.31 &0.51 &5.73 &88.21 \\
& &Motion &11.36 &0.44 &4.63 &15.74 \\
& &Zoom &14.85 &0.38 &4.89 &3.15 \\
\cmidrule(l){2-7}
&\multirow{4}{*}{Weather} &Snow &7.52 &0.39 &5.81 &2.22 \\
& &Frost &8.32 &0.26 &3.26 &45.64 \\
& &Fog &6.54 &0.22 &4.71 &33.75 \\
& &Brightness &7.83 &0.25 &2.84 &32.85 \\
\cmidrule(l){2-7}
&\multirow{4}{*}{Digital} &Contrast &11.41 &0.21 &3.91 &29.58 \\
& &Elastic &10.88 &0.22 &2.74 &27.04 \\
& &Pixelate &9.93 &0.16 &3.85 &0.00 \\
& &JPEG &9.40 &0.19 &2.41 &0.00 \\
\cmidrule(l){2-7}
&\multirow{4}{*}{Geomtric} &Scale &8.26 &0.15 &4.05 &14.88 \\
& &Rotate &5.81 &0.18 &3.96 &15.96 \\
& &Shear &6.61 &0.17 &2.78 &11.21 \\
& &Piecewse Affine &4.48 &0.11 &4.57 &20.07 \\
\end{tabular}
\end{adjustbox}
\begin{adjustbox}{width=0.60\linewidth}
\begin{tabular}{lllrrrrr}
\toprule
\textbf{Watermark} &\textbf{Attack Category} &\textbf{Perturbation} &\textbf{PSNR} &\textbf{SSIM} &\textbf{Bit Acc} &\textbf{Dect Acc} \\
\midrule
\multirow{20}{*}{StegaStamp-WM} &\multirow{4}{*}{Noise} &Gaussian &4.33 &0.12 &1.12 &1.00 \\
& &Shot &4.29 &0.20 &2.04 &1.04 \\
& &Impulse &6.12 &0.14 &2.58 &10.33 \\
& &Speckle &7.87 &0.12 &3.41 &15.89 \\
\cmidrule(l){2-7}
&\multirow{4}{*}{Blur} &Defocus &7.71 &0.18 &5.55 &16.45 \\
& &Glass &12.44 &0.43 &5.89 &12.44 \\
& &Motion &13.28 &0.28 &5.76 &3.33 \\
& &Zoom &13.77 &0.33 &5.04 &1.04 \\
\cmidrule(l){2-7}
&\multirow{4}{*}{Weather} &Snow &1.19 &0.37 &5.94 &5.58 \\
& &Frost &3.21 &0.30 &5.89 &22.89 \\
& &Fog &3.91 &0.29 &3.20 &25.75 \\
& &Brightness &4.22 &0.22 &4.60 &23.06 \\
\cmidrule(l){2-7}
&\multirow{4}{*}{Digital} &Contrast &12.87 &0.24 &1.05 &24.85 \\
& &Elastic &19.30 &0.21 &2.00 &20.00 \\
& &Pixelate &3.34 &0.19 &3.85 &1.41 \\
& &JPEG &5.69 &0.18 &4.58 &1.63 \\
\cmidrule(l){2-7}
&\multirow{4}{*}{Geomtric} &Scale &6.38 &0.16 &3.88 &0.00 \\
& &Rotate &2.48 &0.10 &1.03 &0.00 \\
& &Shear &3.90 &0.15 &2.43 &2.24 \\
& &Piecewse Affine &1.86 &0.15 &3.85 &3.90 \\
\bottomrule
\end{tabular}
\end{adjustbox}
\end{table*}

\begin{table*}[htp]\small
\centering
\caption{LCMs image watermarks under image perturbations.}
\begin{adjustbox}{width=0.60\linewidth}
\begin{tabular}{lllrrrrr}\toprule
\textbf{Watermark} &\textbf{Attack Category} &\textbf{Perturbation} &\textbf{PSNR} &\textbf{SSIM} &\textbf{Bit Acc} &\textbf{Dect Acc} \\\midrule
\multirow{20}{*}{DctDwtSvd-WM} &\multirow{4}{*}{Noise} &Gaussian &18.85 &0.24 &1.53 &19.66 \\
& &Shot &17.81 &0.23 &1.44 &20.34 \\
& &Impulse &18.46 &0.23 &4.51 &18.28 \\
& &Speckle &19.34 &0.31 &2.24 &44.70 \\
\cmidrule(l){2-7}
&\multirow{4}{*}{Blur} &Defocus &29.37 &0.81 &18.77 &39.10 \\
& &Glass &33.14 &0.91 &24.74 &96.58 \\
& &Motion &25.81 &0.78 &12.02 &14.33 \\
& &Zoom &19.29 &0.67 &4.27 &0.00 \\
\cmidrule(l){2-7}
&\multirow{4}{*}{Weather} &Snow &11.13 &0.49 &0.01 &12.71 \\
& &Frost &17.19 &0.73 &7.50 &24.38 \\
& &Fog &14.03 &0.73 &0.72 &1.22 \\
& &Brightness &13.56 &0.83 &0.02 &38.52 \\
\cmidrule(l){2-7}
&\multirow{4}{*}{Digital} &Contrast &16.35 &0.70 &0.97 &3.04 \\
& &Elastic &21.93 &0.64 &11.24 &26.81 \\
& &Pixelate &25.76 &0.69 &13.33 &6.41 \\
& &JPEG &31.18 &0.85 &7.85 &0.00 \\
\cmidrule(l){2-7}
&\multirow{4}{*}{Geomtric} &Scale &13.68 &0.43 &2.07 &0.00 \\
& &Rotate &10.54 &0.33 &1.37 &0.00 \\
& &Shear &13.62 &0.43 &3.41 &0.00 \\
& &Piecewse Affine &18.48 &0.58 &8.48 &10.05 \\
\end{tabular}
\end{adjustbox}
\begin{adjustbox}{width=0.60\linewidth}
\begin{tabular}{lllrrrrr}
\toprule
\textbf{Watermark} &\textbf{Attack Category} &\textbf{Perturbation} &\textbf{PSNR} &\textbf{SSIM} &\textbf{Bit Acc} &\textbf{Dect Acc} \\
\midrule
\multirow{20}{*}{RivaGAN-WM} &\multirow{4}{*}{Noise} &Gaussian &18.78 &0.24 &1.51 &7.39 \\
& &Shot &17.77 &0.22 &1.42 &6.15 \\
& &Impulse &18.41 &0.23 &3.90 &3.88 \\
& &Speckle &19.29 &0.31 &2.09 &13.59 \\
\cmidrule(l){2-7}
&\multirow{4}{*}{Blur} &Defocus &28.94 &0.80 &10.22 &51.71 \\
& &Glass &31.89 &0.88 &10.28 &97.60 \\
& &Motion &25.58 &0.77 &7.46 &52.01 \\
& &Zoom &19.28 &0.66 &3.77 &1.18 \\
\cmidrule(l){2-7}
&\multirow{4}{*}{Weather} &Snow &11.17 &0.48 &0.01 &6.57 \\
& &Frost &15.79 &0.71 &1.31 &43.64 \\
& &Fog &14.04 &0.72 &0.73 &6.07 \\
& &Brightness &13.65 &0.79 &0.04 &52.79 \\
\cmidrule(l){2-7}
&\multirow{4}{*}{Digital} &Contrast &16.34 &0.69 &0.97 &6.07 \\
& &Elastic &21.86 &0.63 &6.27 &91.13 \\
& &Pixelate &25.63 &0.69 &8.44 &9.75 \\
& &JPEG &30.78 &0.84 &7.19 &1.32 \\
\cmidrule(l){2-7}
&\multirow{4}{*}{Geomtric} &Scale &13.68 &0.42 &1.92 &84.81 \\
& &Rotate &10.54 &0.33 &1.29 &16.27 \\
& &Shear &13.63 &0.43 &2.45 &75.16 \\
& &Piecewse Affine &18.46 &0.57 &5.35 &89.71 \\
\end{tabular}
\end{adjustbox}
\begin{adjustbox}{width=0.60\linewidth}
\begin{tabular}{lllrrrrr}
\toprule
\textbf{Watermark} &\textbf{Attack Category} &\textbf{Perturbation} &\textbf{PSNR} &\textbf{SSIM} &\textbf{Bit Acc} &\textbf{Dect Acc} \\
\midrule
\multirow{20}{*}{SSL-WM} &\multirow{4}{*}{Noise} &Gaussian &9.62 &0.38 &1.45 &0.42 \\
& &Shot &7.13 &0.28 &2.40 &0.75 \\
& &Impulse &7.55 &0.75 &3.99 &0.57 \\
& &Speckle &4.46 &0.45 &3.45 &1.52 \\
\cmidrule(l){2-7}
&\multirow{4}{*}{Blur} &Defocus &23.25 &0.38 &4.13 &18.52 \\
& &Glass &20.15 &0.28 &2.25 &3.58 \\
& &Motion &11.52 &0.20 &3.50 &5.23 \\
& &Zoom &10.95 &0.28 &1.55 &6.93 \\
\cmidrule(l){2-7}
&\multirow{4}{*}{Weather} &Snow &4.32 &0.11 &0.66 &8.76 \\
& &Frost &5.48 &0.28 &0.84 &11.03 \\
& &Fog &6.32 &0.48 &1.24 &20.41 \\
& &Brightness &6.12 &0.38 &1.56 &44.21 \\
\cmidrule(l){2-7}
&\multirow{4}{*}{Digital} &Contrast &6.84 &0.50 &3.56 &0.41 \\
& &Elastic &5.85 &0.48 &4.10 &92.45 \\
& &Pixelate &9.16 &0.11 &1.46 &3.31 \\
& &JPEG &8.28 &0.48 &4.50 &2.54 \\
\cmidrule(l){2-7}
&\multirow{4}{*}{Geomtric} &Scale &9.84 &0.33 &3.94 &1.15 \\
& &Rotate &8.48 &0.26 &6.37 &7.88 \\
& &Shear &2.04 &0.24 &4.76 &5.47 \\
& &Piecewse Affine &7.64 &0.15 &5.47 &5.10 \\
\end{tabular}
\end{adjustbox}
\begin{adjustbox}{width=0.60\linewidth}
\begin{tabular}{lllrrrrr}
\toprule
\textbf{Watermark} &\textbf{Attack Category} &\textbf{Perturbation} &\textbf{PSNR} &\textbf{SSIM} &\textbf{Bit Acc} &\textbf{Dect Acc} \\
\midrule
\multirow{20}{*}{StegaStamp-WM} &\multirow{4}{*}{Noise} &Gaussian &2.91 &0.16 &12.84 &18.82 \\
& &Shot &4.92 &0.04 &2.74 &0.94 \\
& &Impulse &5.46 &0.14 &2.52 &2.89 \\
& &Speckle &7.68 &0.13 &3.48 &25.13 \\
\cmidrule(l){2-7}
&\multirow{4}{*}{Blur} &Defocus &8.74 &0.33 &5.34 &22.31 \\
& &Glass &12.66 &0.17 &5.40 &11.38 \\
& &Motion &14.49 &0.10 &6.18 &5.78 \\
& &Zoom &12.15 &0.21 &4.24 &9.45 \\
\cmidrule(l){2-7}
&\multirow{4}{*}{Weather} &Snow &0.49 &0.19 &6.84 &0.21 \\
& &Frost &3.82 &0.28 &4.85 &20.30 \\
& &Fog &3.46 &0.49 &1.20 &25.19 \\
& &Brightness &5.60 &0.04 &3.28 &29.80 \\
\cmidrule(l){2-7}
&\multirow{4}{*}{Digital} &Contrast &13.30 &0.24 &3.45 &25.79 \\
& &Elastic &18.82 &0.30 &0.04 &21.78 \\
& &Pixelate &5.00 &0.18 &4.14 &6.78 \\
& &JPEG &3.92 &0.21 &3.88 &6.82 \\
\cmidrule(l){2-7}
&\multirow{4}{*}{Geomtric} &Scale &6.84 &0.40 &4.94 &0.07 \\
& &Rotate &2.83 &0.21 &1.98 &3.45 \\
& &Shear &4.55 &0.12 &2.72 &11.42 \\
& &Piecewse Affine &1.12 &0.30 &2.88 &3.98 \\
\bottomrule
\end{tabular}
\end{adjustbox}
\end{table*}

\begin{table*}[htp]\small
\centering
\caption{RPG image watermarks under image perturbations.}
\begin{adjustbox}{width=0.60\linewidth}
\begin{tabular}{lllrrrrr}\toprule
\textbf{Watermark} &\textbf{Attack Category} &\textbf{Perturbation} &\textbf{PSNR} &\textbf{SSIM} &\textbf{Bit Acc} &\textbf{Dect Acc} \\\midrule
\multirow{20}{*}{DctDwtSvd-WM} &\multirow{4}{*}{Noise} &Gaussian &21.21 &0.53 &4.42 &38.39 \\
& &Shot &17.95 &0.45 &2.97 &23.52 \\
& &Impulse &21.83 &0.58 &4.74 &39.05 \\
& &Speckle &26.15 &0.73 &6.71 &56.49 \\
\cmidrule(l){2-7}
&\multirow{4}{*}{Blur} &Defocus &17.25 &0.48 &2.72 &17.63 \\
& &Glass &17.17 &0.50 &2.70 &16.35 \\
& &Motion &26.21 &0.78 &6.80 &53.85 \\
& &Zoom &22.06 &0.67 &4.94 &35.21 \\
\cmidrule(l){2-7}
&\multirow{4}{*}{Weather} &Snow &15.77 &0.51 &2.13 &7.47 \\
& &Frost &20.78 &0.67 &4.41 &27.84 \\
& &Fog &15.66 &0.54 &2.12 &5.04 \\
& &Brightness &15.57 &0.55 &2.10 &3.71 \\
\cmidrule(l){2-7}
&\multirow{4}{*}{Digital} &Contrast &19.05 &0.67 &3.69 &17.57 \\
& &Elastic &8.15 &0.37 &1.20 &29.81 \\
& &Pixelate &9.02 &0.41 &0.79 &27.06 \\
& &JPEG &14.79 &0.60 &1.83 &3.48 \\
\cmidrule(l){2-7}
&\multirow{4}{*}{Geomtric} &Scale &22.48 &0.14 &15.02 &25.19 \\
& &Rotate &15.64 &0.29 &1.19 &30.92 \\
& &Shear &14.58 &0.28 &4.05 &55.83 \\
& &Piecewse Affine &13.57 &0.24 &1.29 &69.31 \\
\end{tabular}
\end{adjustbox}
\begin{adjustbox}{width=0.60\linewidth}
\begin{tabular}{lllrrrrr}
\toprule
\textbf{Watermark} &\textbf{Attack Category} &\textbf{Perturbation} &\textbf{PSNR} &\textbf{SSIM} &\textbf{Bit Acc} &\textbf{Dect Acc} \\
\midrule
\multirow{20}{*}{RivaGAN-WM} &\multirow{4}{*}{Noise} &Gaussian &21.33 &0.54 &5.98 &38.89 \\
& &Shot &18.03 &0.46 &3.83 &22.22 \\
& &Impulse &21.95 &0.59 &6.39 &42.90 \\
& &Speckle &26.34 &0.74 &9.24 &65.93 \\
\cmidrule(l){2-7}
&\multirow{4}{*}{Blur} &Defocus &17.32 &0.49 &3.37 &19.77 \\
& &Glass &17.24 &0.50 &3.32 &19.80 \\
& &Motion &26.40 &0.79 &9.28 &67.48 \\
& &Zoom &22.19 &0.68 &6.54 &46.17 \\
\cmidrule(l){2-7}
&\multirow{4}{*}{Weather} &Snow &15.83 &0.51 &2.39 &13.68 \\
& &Frost &20.90 &0.68 &5.70 &40.31 \\
& &Fog &15.71 &0.54 &2.32 &13.89 \\
& &Brightness &15.62 &0.56 &2.26 &13.85 \\
\cmidrule(l){2-7}
&\multirow{4}{*}{Digital} &Contrast &19.15 &0.68 &4.56 &32.48 \\
& &Elastic &8.11 &0.37 &2.63 &24.13 \\
& &Pixelate &8.99 &0.41 &2.06 &19.16 \\
& &JPEG &14.83 &0.60 &1.75 &11.41 \\
\cmidrule(l){2-7}
&\multirow{4}{*}{Geomtric} &Scale &15.72 &0.24 &4.45 &12.42 \\
& &Rotate &12.68 &0.28 &3.72 &27.98 \\
& &Shear &13.66 &0.18 &2.18 &25.54 \\
& &Piecewse Affine &14.03 &0.25 &3.79 &32.88 \\
\end{tabular}
\end{adjustbox}
\begin{adjustbox}{width=0.60\linewidth}
\begin{tabular}{lllrrrrr}
\toprule
\textbf{Watermark} &\textbf{Attack Category} &\textbf{Perturbation} &\textbf{PSNR} &\textbf{SSIM} &\textbf{Bit Acc} &\textbf{Dect Acc} \\
\midrule
\multirow{20}{*}{SSL-WM} &\multirow{4}{*}{Noise} &Gaussian &18.40 &0.27 &3.70 &21.26 \\
& &Shot &13.65 &0.21 &2.81 &11.83 \\
& &Impulse &18.45 &0.32 &3.96 &27.87 \\
& &Speckle &23.86 &0.44 &5.23 &45.50 \\
\cmidrule(l){2-7}
&\multirow{4}{*}{Blur} &Defocus &11.56 &0.23 &2.74 &15.97 \\
& &Glass &11.08 &0.24 &2.76 &17.93 \\
& &Motion &22.80 &0.49 &5.38 &52.36 \\
& &Zoom &16.86 &0.40 &4.24 &39.77 \\
\cmidrule(l){2-7}
&\multirow{4}{*}{Weather} &Snow &8.06 &0.25 &2.49 &19.57 \\
& &Frost &14.39 &0.39 &3.96 &39.65 \\
& &Fog &7.14 &0.28 &2.54 &23.58 \\
& &Brightness &6.65 &0.29 &2.56 &25.49 \\
\cmidrule(l){2-7}
&\multirow{4}{*}{Digital} &Contrast &10.93 &0.39 &3.59 &40.12 \\
& &Elastic &4.05 &0.13 &0.53 &3.47 \\
& &Pixelate &3.26 &0.16 &0.82 &8.80 \\
& &JPEG &4.08 &0.32 &2.51 &31.58 \\
\cmidrule(l){2-7}
&\multirow{4}{*}{Geomtric} &Scale &3.44 &0.33 &5.85 &26.35 \\
& &Rotate &3.85 &0.30 &4.07 &22.81 \\
& &Shear &2.49 &0.18 &4.72 &0.00 \\
& &Piecewse Affine &5.84 &0.29 &3.47 &0.00 \\
\end{tabular}
\end{adjustbox}
\begin{adjustbox}{width=0.60\linewidth}
\begin{tabular}{lllrrrrr}
\toprule
\textbf{Watermark} &\textbf{Attack Category} &\textbf{Perturbation} &\textbf{PSNR} &\textbf{SSIM} &\textbf{Bit Acc} &\textbf{Dect Acc} \\
\midrule
\multirow{20}{*}{StegaStamp-WM} &\multirow{4}{*}{Noise} &Gaussian &5.66 &0.49 &1.89 &0.00 \\
& &Shot &6.00 &0.10 &1.50 &0.00 \\
& &Impulse &6.92 &0.15 &3.45 &0.00 \\
& &Speckle &8.00 &0.19 &4.49 &15.50 \\
\cmidrule(l){2-7}
&\multirow{4}{*}{Blur} &Defocus &7.88 &0.32 &4.41 &19.93 \\
& &Glass &12.30 &0.33 &4.90 &11.98 \\
& &Motion &12.78 &0.38 &5.98 &4.19 \\
& &Zoom &13.42 &0.12 &4.43 &4.28 \\
\cmidrule(l){2-7}
&\multirow{4}{*}{Weather} &Snow &3.42 &0.07 &4.29 &2.42 \\
& &Frost &2.04 &0.42 &4.79 &18.14 \\
& &Fog &3.28 &0.13 &3.48 &2.83 \\
& &Brightness &5.52 &0.41 &4.48 &21.48 \\
\cmidrule(l){2-7}
&\multirow{4}{*}{Digital} &Contrast &10.92 &0.01 &1.12 &22.48 \\
& &Elastic &0.32 &0.28 &2.50 &24.39 \\
& &Pixelate &1.82 &0.44 &3.45 &7.58 \\
& &JPEG &4.18 &0.14 &2.58 &4.92 \\
\cmidrule(l){2-7}
&\multirow{4}{*}{Geomtric} &Scale &4.30 &0.32 &5.29 &2.40 \\
& &Rotate &1.83 &0.24 &2.71 &3.39 \\
& &Shear &4.23 &0.47 &0.41 &5.28 \\
& &Piecewse Affine &1.43 &0.19 &3.15 &3.58 \\
\bottomrule
\end{tabular}
\end{adjustbox}
\end{table*}

\begin{table*}[htp]\small
\centering
\caption{NExT-GPT text watermarks under text perturbations.}
\begin{adjustbox}{width=0.75\linewidth}
\begin{tabular}{lllrrrrr}\toprule
\textbf{Watermark} &\textbf{Attack Category} &\textbf{Perturbation} &\textbf{BLEURT} &\textbf{ROUGE} &\textbf{Bit Acc} &\textbf{Dect Acc} \\\midrule
\multirow{19}{*}{KGW-WM} &\multirow{6}{*}{Character-level} &Keyboard &0.40 &57.96 &65.00 &17.70 \\
& &OCR &0.58 &74.09 &65.22 &17.01 \\
& &Character Insert (CI) &0.42 &58.15 &65.77 &18.37 \\
& &Character Replace (CR) &0.40 &58.15 &64.82 &18.37 \\
& &Character Swap (CS) &0.51 &61.40 &68.17 &19.64 \\
& &Character Delete (CD) &0.47 &58.17 &66.30 &15.09 \\
\cmidrule(l){2-7}
&\multirow{5}{*}{Word-level} &Synonym Replacement (SR) &0.54 &62.54 &68.48 &24.83 \\
& &Word Insertion (WI) &0.53 &76.60 &64.02 &25.87 \\
& &Word Swap (WS) &0.54 &59.88 &67.62 &31.01 \\
& &Word Deletion (WD) &0.50 &72.92 &63.66 &26.51 \\
& &Insert Punctuation (IP) &0.06 &6.99 &53.77 &26.49 \\
\cmidrule(l){2-7}
&\multirow{8}{*}{Sentence-level} &Formal &0.02 &0.46 &50.43 &17.72 \\
& &Casual &0.01 &2.44 &50.28 &18.83 \\
& &Passive &0.07 &7.43 &53.74 &17.07 \\
& &Active &0.07 &8.16 &54.86 &14.88 \\
& &Back Translation &0.79 &77.41 &77.96 &17.03 \\
& &SCPN &0.09 &2.15 &44.82 &11.31 \\
& &Bart &0.00 &3.66 &41.06 &16.74 \\
& &Dipper &0.04 &2.75 &42.78 &17.89 \\
\end{tabular}
\end{adjustbox}
\begin{adjustbox}{width=0.75\linewidth}
\begin{tabular}{lllrrrrr}
\toprule
\textbf{Watermark} &\textbf{Attack Category} &\textbf{Perturbation} &\textbf{BLEURT} &\textbf{ROUGE} &\textbf{Bit Acc} &\textbf{Dect Acc} \\
\midrule
\multirow{19}{*}{KTH-WM} &\multirow{6}{*}{Character-level} &Keyboard &0.39 &63.88 &64.88 &18.67 \\
& &OCR &0.56 &77.77 &65.15 &19.43 \\
& &Character Insert (CI) &0.41 &64.02 &64.79 &19.54 \\
& &Character Replace (CR) &0.39 &64.04 &64.74 &19.60 \\
& &Character Swap (CS) &0.49 &66.66 &66.43 &22.99 \\
& &Character Delete (CD) &0.46 &64.17 &65.72 &17.90 \\
\cmidrule(l){2-7}
&\multirow{5}{*}{Word-level} &Synonym Replacement (SR) &0.44 &53.98 &66.24 &26.73 \\
& &Word Insertion (WI) &0.45 &70.53 &63.42 &28.01 \\
& &Word Swap (WS) &0.47 &56.77 &65.67 &30.01 \\
& &Word Deletion (WD) &0.44 &68.20 &63.13 &27.60 \\
& &Insert Punctuation (IP) &0.17 &4.83 &54.08 &29.04 \\
\cmidrule(l){2-7}
&\multirow{8}{*}{Sentence-level} &Formal &0.05 &0.19 &17.69 &9.08 \\
& &Casual &0.04 &0.52 &14.61 &6.63 \\
& &Passive &0.18 &2.88 &55.40 &25.02 \\
& &Active &0.19 &3.03 &55.40 &23.52 \\
& &Back Translation &0.64 &72.46 &71.33 &23.19 \\
& &SCPN &0.00 &23.62 &20.89 &19.70 \\
& &Bart &0.02 &13.77 &21.75 &18.64 \\
& &Dipper &0.06 &15.84 &22.57 &14.75 \\
\end{tabular}
\end{adjustbox}
\begin{adjustbox}{width=0.75\linewidth}
\begin{tabular}{lllrrrrr}
\toprule
\textbf{Watermark} &\textbf{Attack Category} &\textbf{Perturbation} &\textbf{BLEURT} &\textbf{ROUGE} &\textbf{Bit Acc} &\textbf{Dect Acc} \\
\midrule
\multirow{19}{*}{Blackbox-WM} &\multirow{6}{*}{Character-level} &Keyboard &0.42 &65.78 &81.13 &41.82 \\
& &OCR &0.59 &79.28 &82.75 &50.98 \\
& &Character Insert (CI) &0.44 &66.00 &68.85 &40.73 \\
& &Character Replace (CR) &0.42 &66.00 &81.24 &41.02 \\
& &Character Swap (CS) &0.53 &68.98 &81.33 &42.39 \\
& &Character Delete (CD) &0.48 &66.03 &69.65 &39.87 \\
\cmidrule(l){2-7}
&\multirow{5}{*}{Word-level} &Synonym Replacement (SR) &0.55 &66.35 &67.89 &24.18 \\
& &Word Insertion (WI) &0.54 &82.30 &64.64 &30.04 \\
& &Word Swap (WS) &0.56 &63.71 &67.91 &25.92 \\
& &Word Deletion (WD) &0.52 &78.23 &64.20 &30.57 \\
& &Insert Punctuation (IP) &0.12 &7.78 &54.25 &0.00 \\
\cmidrule(l){2-7}
&\multirow{8}{*}{Sentence-level} &Formal &0.03 &4.88 &54.83 &0.00 \\
& &Casual &0.02 &4.20 &53.68 &0.00 \\
& &Passive &0.08 &7.55 &55.48 &0.00 \\
& &Active &0.13 &8.72 &55.59 &0.00 \\
& &Back Translation &0.74 &69.22 &67.96 &20.70 \\
& &SCPN &0.04 &5.36 &33.85 &21.48 \\
& &Bart &0.03 &4.72 &41.27 &22.34 \\
& &Dipper &0.07 &3.58 &39.27 &20.19 \\
\end{tabular}
\end{adjustbox}
\begin{adjustbox}{width=0.75\linewidth}
\begin{tabular}{lllrrrrr}
\toprule
\textbf{Watermark} &\textbf{Attack Category} &\textbf{Perturbation} &\textbf{BLEURT} &\textbf{ROUGE} &\textbf{Bit Acc} &\textbf{Dect Acc} \\
\midrule
\multirow{19}{*}{Unigram-WM} &\multirow{6}{*}{Character-level} &Keyboard &0.31 &44.98 &55.12 &1.65 \\
& &OCR &0.37 &51.62 &55.08 &1.72 \\
& &Character Insert (CI) &0.31 &45.10 &54.82 &1.67 \\
& &Character Replace (CR) &0.31 &45.10 &55.06 &1.67 \\
& &Character Swap (CS) &0.35 &46.68 &55.21 &1.78 \\
& &Character Delete (CD) &0.34 &45.86 &55.31 &1.82 \\
\cmidrule(l){2-7}
&\multirow{5}{*}{Word-level} &Synonym Replacement (SR) &0.32 &26.71 &53.91 &4.98 \\
& &Word Insertion (WI) &0.33 &50.89 &53.60 &5.20 \\
& &Word Swap (WS) &0.33 &48.00 &54.53 &6.61 \\
& &Word Deletion (WD) &0.34 &55.84 &53.22 &5.42 \\
& &Insert Punctuation (IP) &0.33 &0.48 &55.13 &0.00 \\
\cmidrule(l){2-7}
&\multirow{8}{*}{Sentence-level} &Formal &0.10 &0.22 &14.90 &0.00 \\
& &Casual &0.09 &0.20 &13.21 &0.00 \\
& &Passive &0.30 &0.98 &48.80 &0.00 \\
& &Active &0.31 &0.57 &48.71 &0.00 \\
& &Back Translation &0.35 &36.25 &47.92 &7.05 \\
& &SCPN &0.03 &0.77 &46.37 &1.41 \\
& &Bart &0.06 &0.84 &48.13 &3.32 \\
& &Dipper &0.01 &0.37 &50.88 &2.53 \\
\bottomrule
\end{tabular}
\end{adjustbox}
\end{table*}

\begin{table*}[htp]\small
\centering
\caption{Fuyu-8B text watermarks under text perturbations.}
\begin{adjustbox}{width=0.75\linewidth}
\begin{tabular}{lllrrrrr}\toprule
\textbf{Watermark} &\textbf{Attack Category} &\textbf{Perturbation} &\textbf{BLEURT} &\textbf{ROUGE} &\textbf{Bit Acc} &\textbf{Dect Acc} \\\midrule
\multirow{19}{*}{KGW-WM } &\multirow{6}{*}{Character-level} &Keyboard &0.38 &65.71 &93.15 &36.38 \\
& &OCR &0.53 &79.63 &95.93 &34.20 \\
& &Character Insert (CI) &0.39 &65.90 &68.77 &35.20 \\
& &Character Replace (CR) &0.38 &65.87 &93.84 &35.56 \\
& &Character Swap (CS) &0.49 &69.28 &92.96 &36.06 \\
& &Character Delete (CD) &0.45 &65.97 &69.68 &36.30 \\
\cmidrule(l){2-7}
&\multirow{5}{*}{Word-level} &Synonym Replacement (SR) &0.52 &62.00 &71.67 &34.72 \\
& &Word Insertion (WI) &0.53 &75.65 &67.62 &37.74 \\
& &Word Swap (WS) &0.57 &62.39 &71.87 &39.82 \\
& &Word Deletion (WD) &0.54 &74.70 &69.19 &39.40 \\
& &Insert Punctuation (IP) &0.05 &0.00 &53.86 &50.54 \\
\cmidrule(l){2-7}
&\multirow{8}{*}{Sentence-level} &Formal &0.05 &0.02 &53.30 &48.00 \\
& &Casual &0.03 &2.36 &50.20 &32.90 \\
& &Passive &0.02 &8.67 &55.03 &40.20 \\
& &Active &0.03 &0.21 &55.32 &41.10 \\
& &Back Translation &0.69 &68.60 &70.88 &29.30 \\
& &SCPN &0.06 &0.00 &50.08 &28.53 \\
& &Bart &0.03 &0.06 &50.31 &29.90 \\
& &Dipper &0.00 &0.33 &55.79 &23.35 \\
\end{tabular}
\end{adjustbox}
\begin{adjustbox}{width=0.75\linewidth}
\begin{tabular}{lllrrrrr}
\toprule
\textbf{Watermark} &\textbf{Attack Category} &\textbf{Perturbation} &\textbf{BLEURT} &\textbf{ROUGE} &\textbf{Bit Acc} &\textbf{Dect Acc} \\
\midrule
\multirow{19}{*}{KTH-WM} &\multirow{6}{*}{Character-level} &Keyboard &0.38 &72.63 &85.55 &37.58 \\
& &OCR &0.52 &82.97 &87.04 &36.26 \\
& &Character Insert (CI) &0.39 &72.74 &67.63 &37.76 \\
& &Character Replace (CR) &0.38 &72.74 &85.85 &37.70 \\
& &Character Swap (CS) &0.48 &75.23 &85.89 &38.88 \\
& &Character Delete (CD) &0.45 &72.84 &68.19 &36.64 \\
\cmidrule(l){2-7}
&\multirow{5}{*}{Word-level} &Synonym Replacement (SR) &0.44 &57.01 &68.88 &27.70 \\
& &Word Insertion (WI) &0.46 &72.00 &65.96 &43.46 \\
& &Word Swap (WS) &0.49 &58.95 &69.11 &31.66 \\
& &Word Deletion (WD) &0.46 &71.44 &66.08 &32.18 \\
& &Insert Punctuation (IP) &0.08 &0.02 &53.65 &36.22 \\
\cmidrule(l){2-7}
&\multirow{8}{*}{Sentence-level} &Formal &0.08 &0.02 &54.96 &33.60 \\
& &Casual &0.08 &0.28 &54.94 &32.50 \\
& &Passive &0.06 &2.67 &54.82 &33.20 \\
& &Active &0.07 &0.51 &54.83 &33.10 \\
& &Back Translation &0.63 &67.48 &68.33 &29.60 \\
& &SCPN &0.03 &2.84 &54.89 &28.44 \\
& &Bart &0.07 &2.18 &60.03 &27.45 \\
& &Dipper &0.04 &3.96 &53.83 &22.63 \\
\end{tabular}
\end{adjustbox}
\begin{adjustbox}{width=0.75\linewidth}
\begin{tabular}{lllrrrrr}
\toprule
\textbf{Watermark} &\textbf{Attack Category} &\textbf{Perturbation} &\textbf{BLEURT} &\textbf{ROUGE} &\textbf{Bit Acc} &\textbf{Dect Acc} \\
\midrule
\multirow{19}{*}{Blackbox-WM} &\multirow{6}{*}{Character-level} &Keyboard &0.39 &61.86 &82.67 &49.04 \\
& &OCR &0.53 &74.36 &84.14 &58.80 \\
& &Character Insert (CI) &0.40 &61.96 &69.04 &47.20 \\
& &Character Replace (CR) &0.39 &61.98 &82.58 &47.68 \\
& &Character Swap (CS) &0.49 &65.19 &83.16 &52.24 \\
& &Character Delete (CD) &0.45 &62.07 &69.88 &48.72 \\
\cmidrule(l){2-7}
&\multirow{5}{*}{Word-level} &Synonym Replacement (SR) &0.52 &64.62 &72.19 &45.52 \\
& &Word Insertion (WI) &0.53 &79.48 &68.94 &61.32 \\
& &Word Swap (WS) &0.58 &64.81 &72.34 &38.62 \\
& &Word Deletion (WD) &0.53 &77.35 &69.72 &53.86 \\
& &Insert Punctuation (IP) &0.04 &0.25 &54.28 &0.00 \\
\cmidrule(l){2-7}
&\multirow{8}{*}{Sentence-level} &Formal &0.04 &0.01 &55.96 &0.00 \\
& &Casual &0.03 &2.33 &55.68 &0.00 \\
& &Passive &0.02 &8.23 &55.58 &0.00 \\
& &Active &0.02 &0.12 &55.87 &0.00 \\
& &Back Translation &0.67 &64.88 &68.75 &27.70 \\
& &SCPN &0.01 &3.05 &49.21 &0.00 \\
& &Bart &0.00 &4.66 &50.78 &0.00 \\
& &Dipper &0.04 &2.89 &55.49 &0.00 \\
\end{tabular}
\end{adjustbox}
\begin{adjustbox}{width=0.75\linewidth}
\begin{tabular}{lllrrrrr}
\toprule
\textbf{Watermark} &\textbf{Attack Category} &\textbf{Perturbation} &\textbf{BLEURT} &\textbf{ROUGE} &\textbf{Bit Acc} &\textbf{Dect Acc} \\
\midrule
\multirow{19}{*}{Unigram-WM} &\multirow{6}{*}{Character-level} &Keyboard &0.28 &40.48 &65.05 &2.21 \\
& &OCR &0.42 &53.15 &67.04 &1.10 \\
& &Character Insert (CI) &0.30 &40.56 &49.84 &0.00 \\
& &Character Replace (CR) &0.28 &40.58 &65.04 &0.00 \\
& &Character Swap (CS) &0.37 &43.63 &64.33 &0.00 \\
& &Character Delete (CD) &0.33 &40.62 &50.75 &0.00 \\
\cmidrule(l){2-7}
&\multirow{5}{*}{Word-level} &Synonym Replacement (SR) &0.40 &46.96 &53.28 &4.32 \\
& &Word Insertion (WI) &0.41 &57.98 &49.88 &0.00 \\
& &Word Swap (WS) &0.49 &49.13 &55.10 &0.00 \\
& &Word Deletion (WD) &0.43 &58.53 &53.43 &0.00 \\
& &Insert Punctuation (IP) &0.02 &1.94 &36.91 &2.86 \\
\cmidrule(l){2-7}
&\multirow{8}{*}{Sentence-level} &Formal &0.02 &0.54 &39.36 &0.00 \\
& &Casual &0.02 &3.06 &39.65 &0.00 \\
& &Passive &0.02 &3.75 &39.85 &1.52 \\
& &Active &0.02 &2.93 &39.71 &0.00 \\
& &Back Translation &0.51 &44.35 &54.19 &0.00 \\
& &SCPN &0.01 &3.28 &38.48 &1.49 \\
& &Bart &0.00 &1.13 &33.04 &0.00 \\
& &Dipper &0.15 &2.07 &35.78 &1.13 \\
\bottomrule
\end{tabular}
\end{adjustbox}
\end{table*}

\begin{table*}[htp]\small
\centering
\caption{InternLM-X Composer text watermarks under text perturbations.}
\begin{adjustbox}{width=0.75\linewidth}
%
\end{adjustbox}
\end{table*}

\begin{table*}[htp]\small
\centering
\caption{Instruct-BLIP text watermarks under text perturbations.}
\begin{adjustbox}{width=0.75\linewidth}
%
\end{adjustbox}
\end{table*}

\begin{table*}[htp]\small
\centering
\caption{LLaVA 1.6 text watermarks under text perturbations.}
\begin{adjustbox}{width=0.75\linewidth}
%
\end{adjustbox}
\end{table*}

\begin{table*}[htp]\small
\centering
\caption{MiniGPT-4 text watermarks under text perturbations.}
\begin{adjustbox}{width=0.75\linewidth}
%
\end{adjustbox}
\end{table*}

\begin{table*}[htp]\small
\centering
\caption{mPLUG-Owl2 text watermarks under text perturbations.}
\begin{adjustbox}{width=0.75\linewidth}
%
\end{adjustbox}
\end{table*}

\begin{table*}[htp]\small
\centering
\caption{Qwen-VL text watermarks under text perturbations.}
\begin{adjustbox}{width=0.75\linewidth}
%
\end{adjustbox}
\end{table*}

\end{document}